%% file: main.tex
\documentclass[10pt,twocolumn,letterpaper]{article}

\usepackage[pagenumbers]{iccv} % To force page numbers, e.g. for an arXiv version


\definecolor{iccvblue}{rgb}{0.21,0.49,0.74}
\usepackage[pagebackref,breaklinks,colorlinks,allcolors=iccvblue]{hyperref}
\usepackage{xcolor}
\usepackage{multirow}
\usepackage{multicol}
\usepackage{colortbl}
\usepackage{verbatim}
\usepackage{tcolorbox}

\def\paperID{3372} % *** Enter the Paper ID here
\def\confName{ICCV}
\def\confYear{2025}

\title{ChatRex: Taming Multimodal LLM for Joint Perception and Understanding}

\author{Qing Jiang$^{1,2}$ , Gen Luo$^{1}$ , Yuqin Yang$^{1,2}$ , Yuda Xiong$^{1}$ , Zhaoyang Zeng$^{1}$ \\ Yihao Chen$^{1}$ , Tianhe Ren$^{1}$ , Lei Zhang$^{1,2\dagger}$ \\
$^1$International Digital Economy Academy (IDEA) \\
$^2$South China University of Technology \\ 
{\tt\small mountchicken@outlook.com , leizhang@idea.edu.cn} \\
}

\begin{document}
\maketitle
\makeatletter
\def\@makefnmark{} 
\makeatother
\footnotetext{This work was done when Qing Jiang, Gen Luo and were interns at IDEA.}
\addtocounter{footnote}{-1}
\footnotetext{\emph{$\dagger$ Corresponding author.}}

\input{sec/0_abstract}    
\input{sec/1_intro}
\input{sec/2_Related_work}

\input{sec/3_Method}

\input{sec/4_Data_and_Training}

\input{sec/5_experiment}

\input{sec/6_Conclusion}
{
    \small
    \bibliographystyle{ieeenat_fullname}
    \bibliography{main}
}

% WARNING: do not forget to delete the supplementary pages from your submission 
\input{sec/X_suppl}

\end{document}

%% file: sec/0_abstract.tex
\begin{abstract}
Perception and understanding are two pillars of computer vision. While multimodal large language models (MLLM) have demonstrated remarkable visual understanding capabilities, they arguably lack accurate perception abilities, e.g. the stage-of-the-art model Qwen2-VL only achieves a 43.9 recall rate on the COCO dataset, limiting many tasks requiring the combination of perception and understanding. In this work, we aim to bridge this perception gap from both model designing and data development perspectives. We first introduce ChatRex, an MLLM with a decoupled perception design. Instead of having the LLM directly predict box coordinates, we feed the output boxes from a universal proposal network into the LLM, allowing it to output the corresponding box indices to represent its detection results, turning the regression task into a retrieval-based task that LLM handles more proficiently. From the data perspective, we build a fully automated data engine and construct the Rexverse-2M dataset which possesses multiple granularities to support the joint training of perception and understanding. After a three-stage training approach, ChatRex demonstrates strong perception and understanding performance, and the combination of these two capabilities also unlocks many attractive applications, demonstrating their complementary roles in MLLM. Code is available at \url{https://github.com/IDEA-Research/ChatRex}.

\end{abstract}

%% file: sec/1_intro.tex
\section{Introduction}
\label{sec:intro}

\begin{figure*}[h]\centering
\includegraphics[width=0.92\linewidth]{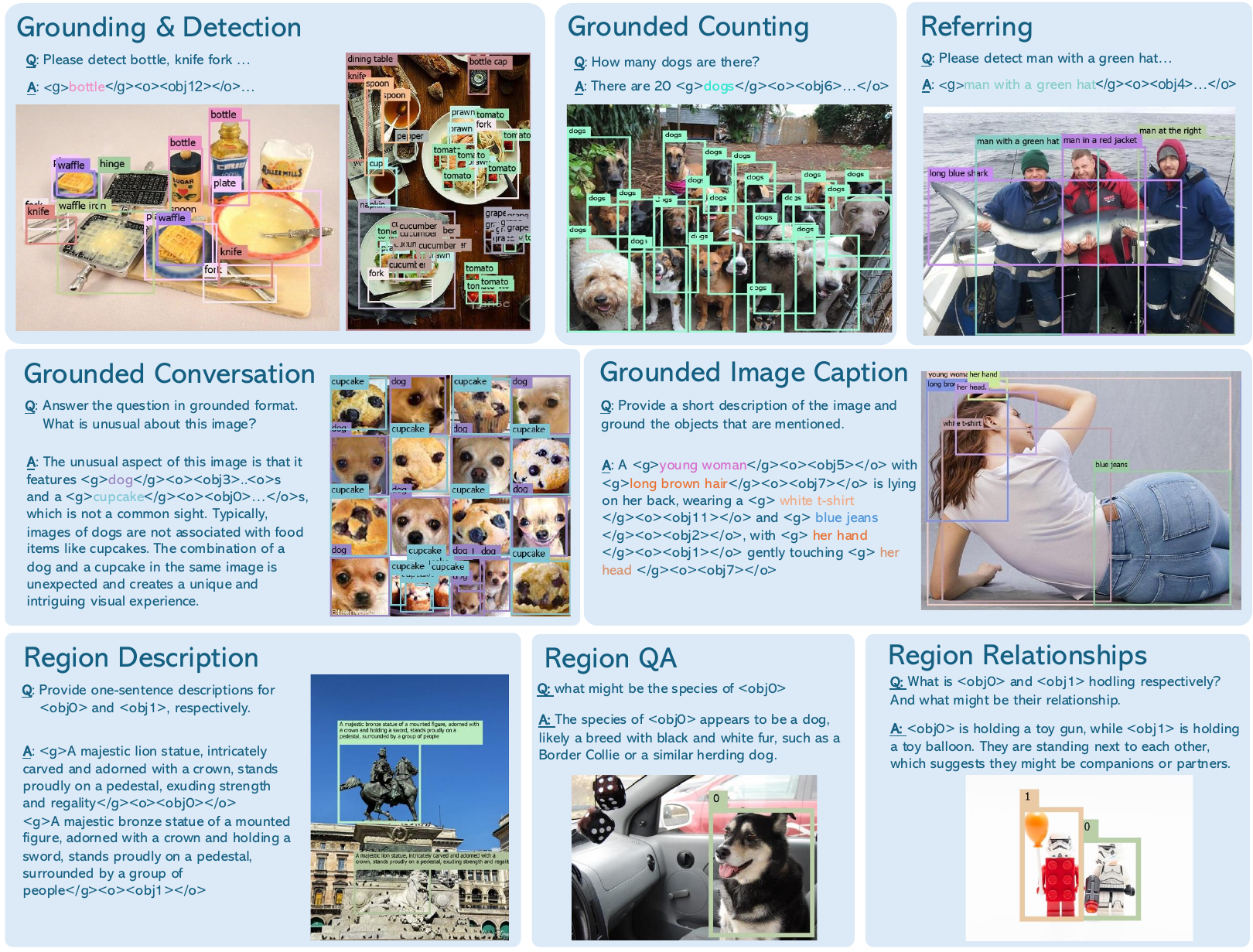}\vspace{-1mm}
\caption{Overview of the perception capabilities in ChatRex. We utilize a decoupled design for perception and understanding, allowing ChatRex to respond to questions while simultaneously grounding its answers to the referenced objects.}
\label{fig:teaser}
\vspace{-1mm}
\end{figure*}

Perception and understanding are two fundamental human faculties within behavioral science. Humans initially perceive objects, with vision signals transmitted to the brain for understanding, and can then locate back to the objects during conversation. In pursuit of AGI, Multimodal Large Language Models (MLLMs)~\cite{gpt4v, VLM:InternVL-1.5, VLM:Gemini, bai2023qwenvl, alayrac2022flamingo, chen2022pali, VLM:InstructBLIP, agrawal2024pixtral, li2024aria, deitke2024molmo, wang2024qwen2, lv2023kosmos2_5} have exhibited remarkable capacities for visual understanding empowered by advancements in Large Language Models (LLMs)~\cite{TransF:LLaMA, TransF:LLaMA2, TransF:Vicuna, TransF:Alpaca, TransF:InternLM, openai2023gpt4, qwen, abdin2024phi, dubey2024llama}. Despite showing strong visual understanding, we find through experiments that these models generally lack fine-grained perception capabilities, particularly in object detection.

We evaluate the performance of several general-purpose~\cite{VLM:InternVL-1.5, wang2024qwen2} and detection-focused~\cite{you2023ferret, chen2023shikra, ma2024groma} MLLMs on the COCO~\cite{Datasets:MSCOCO} dataset by prompting them to detect objects within the image. The state-of-the-art model Qwen2-VL-7B ~\cite{wang2024qwen2} only achieves a 43.9 recall rate at an IoU threshold of 0.5. The results indicate that MLLMs still struggle with fundamental perception tasks despite their remarkable visual understanding capabilities. This shortfall in perception constrains them in numerous tasks requiring precise perception, such as autonomous driving and robotic navigation. Also, it hinders their interactivity by identifying objects during conversation. We argue that this performance gap between perception and understanding in MLLMs arises primarily from two factors: \textit{i)} modeling conflicts between these two tasks, and \textit{ii)} lack of data that seamlessly balances both perception and understanding.

For object detection, a common practice is to quantize~\cite{chen2021pix2seq} box coordinates into tokens within the vocabulary of LLM to fit the auto-regressive framework. Although this ensures compatibility with understanding tasks through next-token prediction, we argue this method is in conflict with accurately modeling perception for three reasons: \textit{i)} Error propagation: representing a single box typically requires 9 tokens including digits, square brackets, and commas, where an error in any token can cause cascading errors, which become even worse in multi-object detection. In subsequent experiments, we find that this is one of the reasons for the low recall rate; \textit{ii)} Ambiguity in prediction order: there is no inherent order among objects in object perception, yet the auto-regressive nature imposes a sequential order that the LLM must decide which object to predict first and \textit{iii)} Quantization range limitation: quantization error easily occurs when the image size is large.

To address these inherent modeling conflicts, we adopt a decoupled model design and introduce ChatRex. For multimodal understanding tasks like image caption and image QA, we retain the auto-regressive text prediction framework. However, for perception, particularly object detection, we transform the task as a retrieval-based task inspired by Groma~\cite{ma2024groma}. Specifically, instead of prompting the LLM to predict bounding box coordinates, the boxes are directly provided as inputs, each represented as an object token by combining its RoI feature with its positional embedding. When the LLM needs to reference an object, it outputs the index of the relevant box. This method represents each box as a single token without quantization, with the sequence order determined by the input boxes, effectively addressing prior modeling conflicts. 

However, this retrieval-based approach presents two key challenges for achieving optimal performance: the need for high-resolution visual input and a robust object proposal model. To address the first challenge, we adopt a dual vision encoder design to incorporate additional vision encoder~\cite{liu2022convnet} to provide high-resolution visual information for perception. For the second, we introduce a Universal Proposal Network (UPN), which leverages granularity-based prompt learning on a pre-trained open-set object detection model. This enables the generation of proposals that cover diverse granularities, categories, and domains, thereby ensuring robust box inputs for the LLM.

From the data perspective, current MLLMs are also limited by the lack of data that effectively balances both perception and understanding. To address this limitation, we developed a fully automated data engine to construct the Rexverse-2M dataset, which comprises image-region-text annotation triplets at varying levels of granularity. The data engine is composed of three primary modules. The first module generates image captions for input images~\cite{VLM:InternVL-1.5}, while the second aligns referenced objects or phrases using a grounding model~\cite{ren2024grounding}. The third module~\cite{dubey2024llama} refines region descriptions at multiple granularities.

Experimental results show that ChatRex achieves strong performance in object detection tasks, including COCO\cite{Datasets:MSCOCO}, LVIS\cite{gupta2019lvis}, and RefCOCO/+/g~\cite{kazemzadeh2014referitgame, yu2016refcoco, mao2016refcocog}, while also demonstrating competitive performance on multimodal benchmarks. Our findings highlight that both perception and understanding are fundamental capabilities for multimodal models, and their integration leads to performance gain with expansion on the scope of real-world applications, as illustrated in Fig. \ref{fig:teaser}. To summarize, our contributions are threefold:

\begin{itemize}
    \item We reveal the performance gap in the perception of MLLMs and introduce a decoupled model ChatRex and a universal proposal network (UPN) to address the modeling conflict between perception and understanding.
    \item We develop an automated data engine to create Rexverse-2M, a comprehensive dataset supporting both perception and understanding tasks for model training.
    \item Experimental results demonstrate that ChatRex exhibits strong perception and multimodal understanding capabilities, highlighting that these two complementary abilities are both essential for MLLM.
\end{itemize}

%% file: sec/2_Related_work.tex
\section{Related Work}
\label{sec:related_works}
\subsection{General MLLMs}

Leveraging breakthroughs in large language models within natural language processing, Multimodal Large Language Models (MLLMs)~\cite{gpt4v, VLM:InternVL-1.5, VLM:Gemini, bai2023qwenvl, alayrac2022flamingo, chen2022pali, VLM:InstructBLIP, agrawal2024pixtral, li2024aria, deitke2024molmo, wang2024qwen2, lv2023kosmos2_5} have demonstrated robust visual understanding capabilities. LLaVA~\cite{VLM:LLaVA} pioneered the paradigm of visual instruction tuning, inspiring a wave of subsequent work. Research on general-purpose MLLMs encompasses various directions, including: \textit{i)} exploring the use of high-resolution image inputs to enhance model perceptual abilities, with models like LLaVA-Next~\cite{VLM:LLaVA-1.6}, SPHINX~\cite{lin2023sphinx}, Monkey~\cite{li2023monkey}, InternLM-XComposer2~\cite{VLM:Xcomposer2-4KHD}, LLaVA-UHD~\cite{VLM:LLaVA-UHD}, NVLM~\cite{dai2024nvlm} employing image slicing methods, and others like LLaVA-HR~\cite{luo2024llava_hr}, Mini-Gemini~\cite{VLM:MiniGemini}, Eagle~\cite{shi2024eagle}, and MG-LLaVA~\cite{zhao2024mg} utilizing high-resolution vision encoders for additional vision encoding; \textit{ii)} investigating diverse approaches for pre-training~\cite{lin2024vila, VLM:MM1, xue2024xgen} and fine-tuning data~\cite{li2024llava, tong2024cambrian}, and \textit{iii)} extending to multi image~\cite{VLM:MANTIS, li2024llava} or video tasks~\cite{xue2024longvila, lin2023video}.

\subsection{Perception MLLMs}
While generic multimodal models demonstrate strong image-level understanding, they still lack fine-grained perception capabilities. Inspired by Pix2seq~\cite{chen2021pix2seq}, several works such as Kosmos-2~\cite{peng2023kosmos2}, Shikra~\cite{chen2023shikra}, Ferret~\cite{you2023ferret, zhang2024ferret}, CogVLM~\cite{wang2023cogvlm}, Griffon~\cite{zhan2025griffon, zhan2024griffon} and other generalized MLLMs~\cite{wang2024qwen2, VLM:InternVL-1.5, VLM:MM1} have transformed box regression into a quantized coordinate prediction task suited for LLM next-token prediction. SoM~\cite{yang2023set} uses a set of marks to prompt GPT4V for visual grounding tasks. Another research direction employs additional decoders for perception. For instance, LISA~\cite{lai2023lisa}, GLaMM~\cite{rasheed2024glamm}, LLaVA-Grounding~\cite{zhang2025llava}, PerceptionGPT~\cite{pi2024perceptiongpt}, and VisionLLMv2~\cite{wu2024visionllm} use auxiliary detection or segmentation models for perception tasks. Groma~\cite{ma2024groma} initially proposed re-framing detection as a box retrieval task, and we follow this method in this work.

%% file: sec/3_Method.tex
\section{ChatRex Architecture}
\label{sec:method}

ChatRex employs a design that decouples perception from understanding. For perception, we train a universal proposal network to detect arbitrary objects, supplying box inputs to the LLM. For understanding, we adopt the standard LLaVA~\cite{VLM:LLaVA} structure with a dual vision encoder to facilitate high-resolution image encoding. We introduce each part in the following sections.

\begin{figure}[t]
\centering
\includegraphics[width=0.95\linewidth]{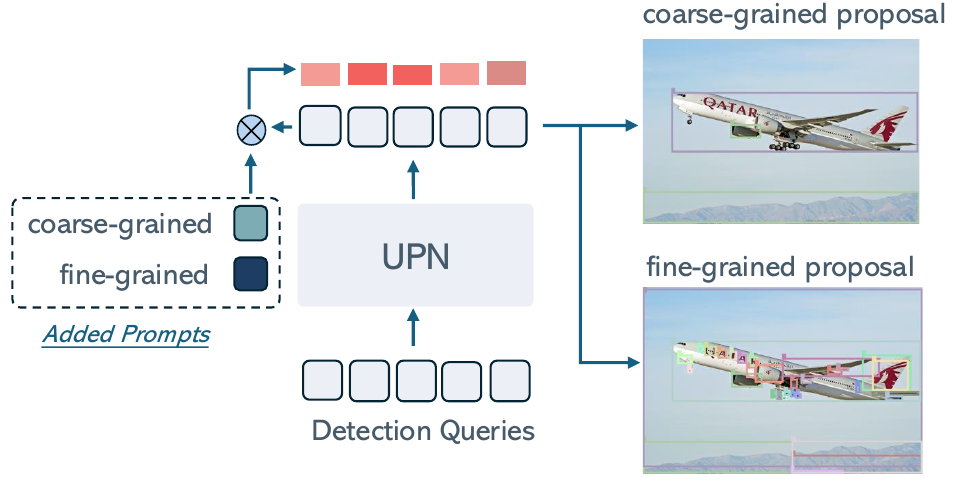}
\caption{Overview structure of the Universal Proposal Network (UPN). UPN is a DETR-based model capable of detecting any object at two granularities.}
\label{fig:proposal} 
\vspace{-3mm}
\end{figure}

\begin{figure*}[h]\centering
\includegraphics[width=0.8\linewidth]{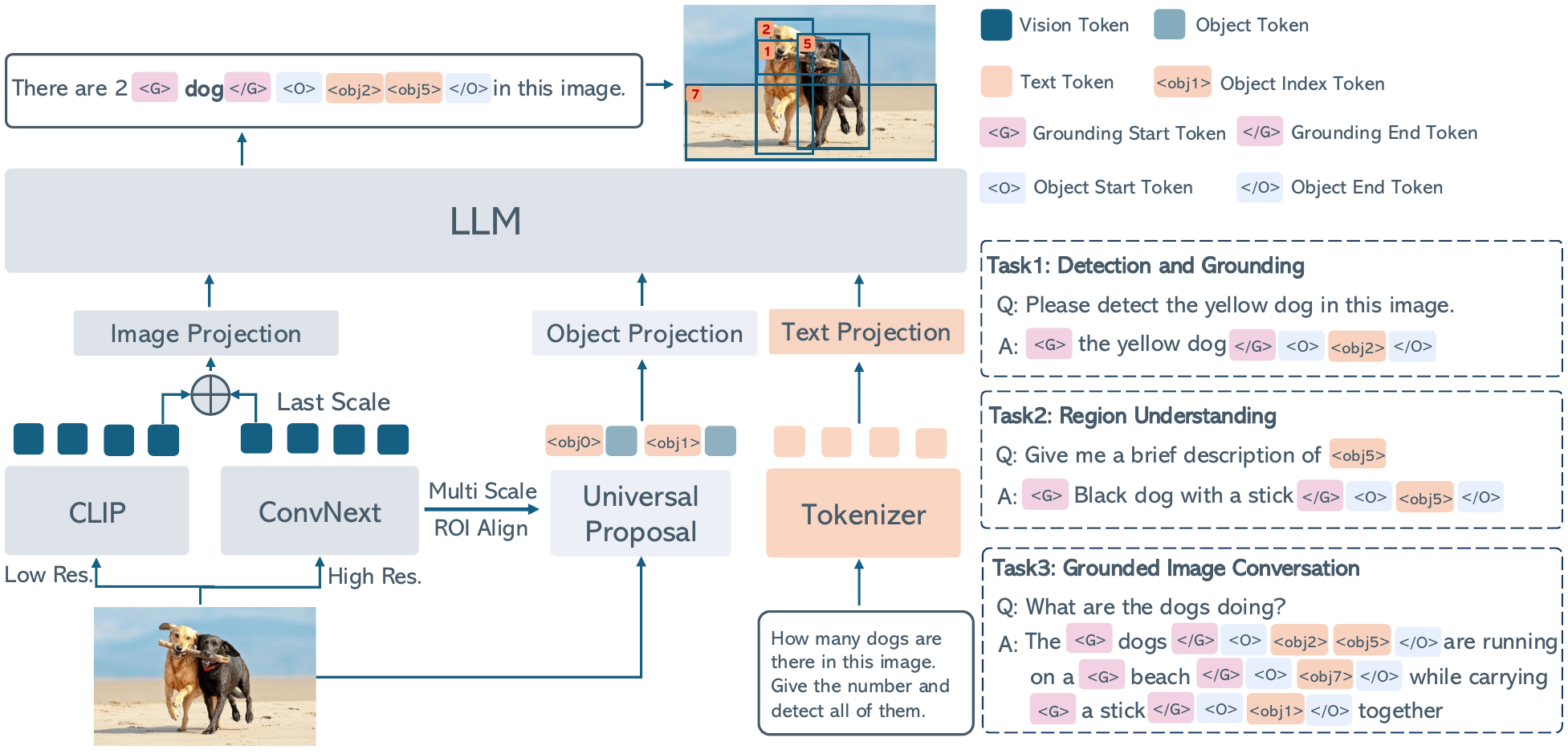}\vspace{-1mm}
\caption{Overview of the proposed ChatRex model architecture and the workflow for modeling the detection output of the LLM from coordinates prediction task to input box indices retrieval task.}
\label{fig:chatrexmodel}
\vspace{-1mm}
\end{figure*}

\subsection{Universal Proposal Network (UPN)}
To ensure that the LLM can accurately retrieve the correct box, it is essential that the input boxes comprehensively encompass all objects within an image. This requires a proposal model with two key properties: \textit{i)} robust generalization ability to generate proposal boxes for any object in any scenario, and \textit{ii)} the proposed boxes should be comprehensive, including both instance-level and part-level objects.

To meet these requirements, a straightforward approach is to aggregate multiple detection datasets, merge their categories, and treat all object classes as a single foreground category for training. However, this strategy is suboptimal due to inconsistencies in object definitions across different datasets. For example, while datasets such as COCO~\cite{Datasets:MSCOCO} and O365~\cite{shao2019objects365} annotate objects at the instance level, SA-1B~\cite{TransF:SAM} annotate objects at part-level. These discrepancies in labeling can introduce ambiguities that compromise training stability. To mitigate this issue, we adopt a dual-granularity prompt tuning strategy.

Specifically, we utilize T-Rex2~\cite{jiang2025t} as our base model. T-Rex2 is a DETR-based~\cite{carion2020end} model trained on vast data and exhibits strong generalization, making it a suitable pre-trained model for detecting any objects in varied scenes. The model outputs object queries \( \mathbf{Q}_{\text{dec}} \) that pass through an MLP to predict bounding boxes. The classification of these bounding boxes is achieved via a dot product between the queries and the prompt embeddings $\mathbf{E}$:
\begin{equation}
\mathbf{S}_{\text{cls}} = \mathbf{E} \cdot \mathbf{Q}_{\text{dec}}^T : \mathbb{R}^{C \times D} \times \mathbb{R}^{D \times N} \rightarrow \mathbb{R}^{C \times N}
\end{equation}
Where $C$ is the number of classes, $N$ represents the number of detection queries (default is 900), and $D$ is the channel dimension of outputted queries. We extend T-Rex2 by introducing two additional learnable prompts, \( \mathbf{P}_{\text{fine}} \) and \( \mathbf{P}_{\text{coarse}} \), concatenated into \( \mathbf{P}_{\text{concat}} \) to classify boxes into fine-grained or coarse-grained categories:
\begin{equation}
\mathbf{S}_{\text{cls}} = \mathbf{P}_{\text{concat}} \cdot \mathbf{Q}_{\text{dec}}^T : \mathbb{R}^{2 \times D} \times \mathbb{R}^{D \times N} \rightarrow \mathbb{R}^{2 \times N}
\end{equation}
For training, we utilize SA-1B as the fine-grained dataset and other detection datasets (such as COCO and O365) as coarse-grained inputs. This dual-granularity prompt design effectively resolves labeling ambiguities between datasets, allowing the proposal model to accurately capture and characterize objects across varying levels of detail.

\subsection{MLLM Architecture}
\textbf{Dual Vision Encoders.}
An accurate perception system typically requires high-resolution inputs. To equip ChatRex with sufficient perception capabilities, we adopt an additional high-resolution vision encoder for image encoding. As illustrated in Fig. \ref{fig:chatrexmodel}, we use the ViT~\cite{TransF:ViT} from CLIP~\cite{VLP:CLIP} for low-resolution image encoding and ConvNeXt~\cite{liu2022convnet} from LAION~\cite{Datasets:Laion-5b} for high-resolution image encoding. To reduce the number of vision tokens fed into the LLM, we first adjust the input resolutions for both vision encoders to ensure they generate the same number of tokens at the last scale. We then directly concatenate these two tokens along the channel dimension, producing the same number of the low-resolution token count.

\begin{figure*}[h]\centering
\includegraphics[width=0.85\linewidth]{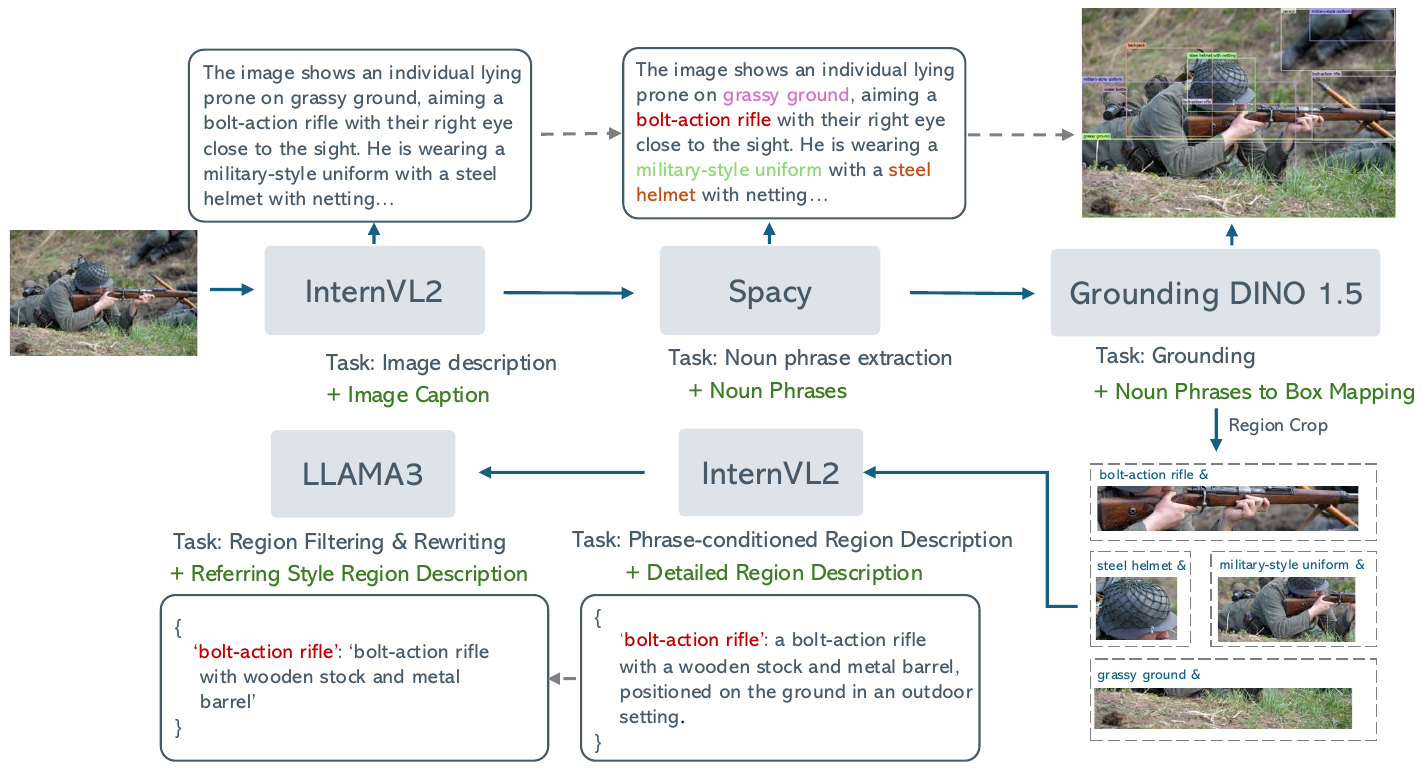}\vspace{-1mm}
\caption{Overview of the ChatRex data engine. There are three main components, including the image captioning module, the grounding module, and the region captioning module.}
\label{fig:data_engine}
\vspace{-1mm}
\end{figure*}

\noindent\textbf{Object Encoder}.
We encode each output box from the universal proposal network to object tokens and feed them to the LLM. Assume $K$ input boxes $\{B_i\}_{i=1}^K$ from the UPN, let \(\mathcal{F}_{\mathrm{H}}\) denote the multi-scale visual features produced by the high-resolution encoder, for each box $B_i$, we extract its content feature $\mathcal{C}_i$ using multi-scale RoI Align~\cite{he2017mask}: 
\begin{equation}
\mathcal{C}_i = \text{RoIAlign}(\mathcal{F}_{\mathrm{H}}, B_i)
\end{equation}
Since the RoI feature does not contain positional information which is essential for referring tasks, we enhance each object feature with a positional embedding to capture spatial context. We encode each box coordinate through a sin-cos position embedding layer and add to the RoI feature:
\begin{equation}
\mathcal{V}_i = \mathcal{C}_i + \operatorname{PE}(B_i)
\end{equation}
\noindent\textbf{LLM}.
We use two separate MLP projectors to map visual and object tokens to the text space. We also add an index token to each object token to inform the LLM about the index of each object token, which will be described in Sec. \ref{task_formulation}. These tokens are then concatenated with the text token and fed into the LLM for the next-token prediction task. We use Qwen2.5-7B~\cite{yang2024qwen2} as our default LLM.

\subsection{Task Formulations}
\label{task_formulation}
We formulate the task of leveraging LLM for detection as an index selection process over input boxes. To do so, we first extend the vocabulary of LLM by incorporating specialized tokens, including object index tokens \texttt{<obj0>}, \texttt{<obj1>}, ..., \texttt{<objN>}, where 
\( N \) denotes the maximum number of input boxes and is set to 100 in this work; grounding start token \texttt{<g>}; grounding end token \texttt{</g>}; object start token \texttt{<o>}; and object end token \texttt{</o>}.

\textbf{LLM Input Format}. The input token sequence for the LLM are structured as follows:
\[
\small
\texttt{<image>\textbackslash n<obj1><roi>\ldots<objN><roi>\textbackslash nQuestion}
\]
where \(\texttt{<image>}\) represents the visual tokens from the vision encoder, and \(\texttt{<roi>}\) denotes the object feature associated with each corresponding bounding box. Each \(\texttt{<roi>}\) token is prefixed by its respective object index token.

\textbf{Decoupled Task Formulations}.
The detection results produced by the LLM are structured using the following combination of noun phrases and box indices:
\[
\small
\texttt{<g>noun phrases</g><o><objm>...<objn><o>}
\]

where \(\texttt{<objm>}\) and \(\texttt{<objn>}\) refer to specific object index tokens, identifying the start (\(m\)) and end (\(n\)) of the sequence of detected objects linked to the noun phrases. This structured format enables a precise mapping between noun phrases and their corresponding bounding box indices.

With this input-output schema, ChatRex can handle various tasks such as detection, grounding, region comprehension, and grounded dialogue, in addition to generating plain text responses, as illustrated in Fig. \ref{fig:chatrexmodel}.

%% file: sec/4_Data_and_Training.tex
\section{Data and Training}
\label{sec:data_and_training}

To equip ChatRex with robust perception and understanding capabilities, we build RexVerse-2M dataset with two million annotated images, featuring multi-granularity annotations generated through a fully automatic data engine. We then adopt a standard two-stage training methodology following LLaVA~\cite{VLM:LLaVA}, enabling the model to preserve its perception capabilities while progressively acquiring multimodal understanding and dialog skills.

\subsection{RexVerse-2M Data Engine}

Our objective is to construct a dataset that can be effectively utilized for both perception and understanding tasks. To achieve this, our data pipeline focuses on generating an annotation triplet comprising image descriptions, region descriptions, and bounding boxes. As shown in Fig. \ref{fig:data_engine}, the data engine is structured around three core modules: image captioning, object grounding, and region captioning.

\textbf{Image Collection.}
We started by collecting images from COYO700M~\cite{kakaobrain2022coyo-700m} dataset through a series of filtering processes including removing images with small resolution and NSFW tags. We also train an image classifier to filter out low-content web images with plain white backgrounds. Finally, we selected two million images as the dataset images.

\textbf{Image Caption.} We use InternVL2-8B~\cite{VLM:InternVL-1.5} to generate image caption for each image. This image caption will refer to the main objects in the image by their category name or descriptive phrases.

\textbf{Phrase Grounding}. We then utilize SpaCy\footnote{\url{https://spacy.io/}} to extract noun phrases from generated image captions. Depending on the caption, SpaCy may identify category names, such as \texttt{soldier} or descriptive phrases (at least 3 words per region) like \texttt{military-style uniform}. We will also filter out some abstract nouns that might not be an object like \texttt{image, background} etc. Subsequently, we employ Grounding DINO 1.5~\cite{ren2024grounding} to ground the filtered noun phrases. This process ultimately produces boxes associated with their category names or short phrase descriptions.

\textbf{Phrase-Conditioned Region Caption}. To support the training for understanding tasks, it is essential to generate detailed descriptions for each region rather than relying solely on category names or short phrases, which often provide limited information. A straightforward approach might involve cropping each region and feeding it into an MLLM model for image captioning. However, this method is prone to hallucinations when the cropped regions are too small or contain parts of other objects. To reduce such inaccuracies, we implemented a phrase-conditioned image description strategy. Specifically, we leverage the InternVL2-8B model~\cite{VLM:InternVL-1.5} to generate image captions that are conditioned on predefined phrases related to each region. By guiding the model with these phrases, we ensure that the generated descriptions are more accurate and context-relevant, significantly reducing the likelihood of hallucinations and enhancing the quality of the region-specific captions.

\textbf{Region Caption Filtering and Rewriting}. Lastly, we employ LLaMA3-8B~\cite{dubey2024llama} to verify whether the generated captions accurately align with their original category names or short phrases, filtering out any remaining hallucinated outputs. Once validated, we then prompt it to refine these detailed captions into more concise referring expressions, thereby enhancing training for referring tasks.

Rexverse-2M consists of 2.1 million images with captions, 10.2 million regions annotated with category labels, 2.5 million regions labeled with short phrases, 2.5 million regions with detailed descriptions, and 2.4 million regions with referring descriptions. Additionally, we use this data engine to annotate 776K grounded conversation data from the ALLaVA-4V-Instruct~\cite{chen2024allava} dataset for instruction tuning. Specifically, the conversation responses are treated as image captions, which are then passed through the engine.

\subsection{Training}
\textbf{UPN Training}.
We utilize two types of datasets with bounding boxes to train our UPN: coarse-grained datasets including O365~\cite{shao2019objects365}, OpenImages~\cite{openimages}, Bamboo~\cite{zhang2022bamboo},  COCO~\cite{Datasets:MSCOCO}, LVIS~\cite{gupta2019lvis}, HierText~\cite{hiertext}, CrowdHuman~\cite{shao2018crowdhuman}, SROIE~\cite{sroie} and EgoObjects~\cite{zhu2023egoobjects}; and fine-grained datasets SA-1B~\cite{TransF:SAM}. All dataset categories are defined as either coarse-grained or fine-grained, reducing the task to a binary classification problem. Following T-Rex2, Hungarian matching is used to match predictions with ground truth. We employ L1 Loss and GIOU Loss for box predictions, along with sigmoid focal loss for classification.

\textbf{ChatRex Training Tasks}.
We adopt three main tasks to train ChatRex including \textit{i)} Grounding: where the model outputs the indices of corresponding objects based on a given category name, phrase, or referring expression. \textit{ii)} Region Understanding: where, given region indices, the model generates descriptions at varying levels of detail, including category names, short phrases, detailed descriptions, or referring descriptions, and \textit{iii)} Grounded Image Conversation: The model needs to output indices of objects mentioned in its generated conversation output. We mix the ground truth boxes of the current image with the proposal boxes from UPN, and keep at most 100 boxes as input. We adopt a three-stage training process and the data for each stage are listed in Tab. \ref{tab:training_data}

\textbf{Stage-1: Alignment Training}.
In the first stage, the objective is to align visual features and object features with the text feature space. To achieve this, we train the image projection MLP, object projection MLP, as well as the input and output embeddings of the LLM, given that we have added special tokens to its vocabulary.

\textbf{Stage-2: Perception Training}. 
In this stage, we improve the perception capability of ChatRex by training it on Rexverse-2M and other grounding data. All the parameters are trainable in this stage.

\textbf{Stage-3: Joint Training}. 
In this stage, we integrate perception and understanding tasks into a unified training process, ensuring that ChatRex acquires both capabilities. This joint optimization equips the model with comprehensive multimodal abilities and enables mutual enhancement between perception and understanding.

\begin{table}[]
  \centering
  \resizebox{1\columnwidth}{!}{%
   \begin{tabular}{c|c|c|c}
\hline
Stage  & Task                                                                                                                                    & \# Samples & Datasets                                                                                                                                                                          \\ \hline
Stage1 & Image Caption                                                                                                                           & 976K       & ALLAVA-4V-Caption~\cite{chen2024allava}                                                                                                                                                                 \\ \hline
Stage2 & \begin{tabular}[c]{@{}c@{}}Grounding \&\\  Region Understanding\end{tabular}                                                            & 2.07M      & \begin{tabular}[c]{@{}c@{}}COCO~\cite{Datasets:MSCOCO}, O365~\cite{shao2019objects365}, LVIS~\cite{gupta2019lvis},\\ RefCOCO/+/g~\cite{kazemzadeh2014referitgame, yu2016refcoco, mao2016refcocog}, Rexverse-2M\end{tabular}                                                                                                           \\ \hline
Stage3 & \begin{tabular}[c]{@{}c@{}}Grounding \& Counting \&\\ Region Understanding \&\\ Grounded Conversation\\ Conversation \& QA\end{tabular} & 3.8M       & \begin{tabular}[c]{@{}c@{}}Rexverse-2M,\\ COCO, O365, LVIS,\\ RefCOCO/+/g~\cite{kazemzadeh2014referitgame, yu2016refcoco, mao2016refcocog}, PACO~\cite{ramanathan2023paco}\\  MVDP~\cite{lin2024draw}, Osprey~\cite{yuan2024osprey}, CrowdHuman~\cite{shao2018crowdhuman},\\ VCR~\cite{zellers2019vcr} , ALLAVA-4V-Instruct~\cite{chen2024allava} ,\\ LLAVA-1.5~\cite{VLM:LLaVA-1.5}, LLaVA-Onevision~\cite{li2024llava}\end{tabular} \\ \hline
\end{tabular}
}
\caption{Training data and tasks for each stage.}
\label{tab:training_data}
\vspace{-1.5em}
\end{table}

%% file: sec/5_experiment.tex
\begin{table*}[t]
\centering
  \resizebox{\linewidth}{!}{
    \begin{tabular}{c|c|ccc|cccccc|ccc|ccc|cc}
\toprule
Method            & Type                                                                                   & \multicolumn{3}{c|}{COCO-Val}                 & \multicolumn{6}{c|}{LVIS-Mini Val}                                                                                                                     & \multicolumn{3}{c|}{RefCOCO}                  & \multicolumn{3}{c|}{RefCOCO+}                 & \multicolumn{2}{c}{RefCOCOg}  \\ \hline
                  &                                                                                        & P@0.5         & R@0.5         & mAP           & \multicolumn{1}{l}{P@0.5} & \multicolumn{1}{l}{R@0.5} & mAP                   & AP-R                  & AP-C                  & AP-F                   & val           & testA         & testB         & val           & testA         & testB         & val           & test          \\ \hline
Faster-RCNN~\cite{Detection:FasterR-CNN}       & \multirow{4}{*}{\begin{tabular}[c]{@{}c@{}}Closed-set \\ Detection Model\end{tabular}} & -             & -             & 42.0          & \multicolumn{1}{l}{-}     & \multicolumn{1}{l}{-}     & \multicolumn{1}{l}{-} & \multicolumn{1}{l}{-} & \multicolumn{1}{l}{-} & \multicolumn{1}{l|}{-} & -             & -             & -             & -             & -             & -             & -             & -             \\
DETR~\cite{carion2020end}              &                                                                                        & -             & -             & 43.3          & \multicolumn{1}{l}{-}     & \multicolumn{1}{l}{-}     & \multicolumn{1}{l}{-} & \multicolumn{1}{l}{-} & \multicolumn{1}{l}{-} & \multicolumn{1}{l|}{-} & -             & -             & -             & -             & -             & -             & -             & -             \\
Pix2Seq~\cite{chen2021pix2seq}           &                                                                                        & -             & -             & 43.2          & \multicolumn{1}{l}{-}     & \multicolumn{1}{l}{-}     & -                     & -                     & -                     & -                      & -             & -             & -             & -             & -             & -             & -             & -             \\
DINO~\cite{zhang2022dino}              &                                                                                        & -             & -             & 49.4          & -                         & -                         & -                     & -                     & -                     & -                      & -             & -             & -             & -             & -             & -             & -             & -             \\ \midrule
Florence2~\cite{xiao2024florence}         & \multirow{4}{*}{\begin{tabular}[c]{@{}c@{}}Open-set \\ Detection Model\end{tabular}}   & -             & -             & 43.4          & -                         & -                         & -                     & -                     & -                     & -                      & -             & -             & -             & -             & -             & -             & -             & -             \\
GLIP~\cite{li2022grounded}              &                                                                                        & -             & -             & \textbf{49.8} & -                         & -                         & 37.3                  & 28.2                  & 34.3                  & 41.5                   & -             & -             & -             & -             & -             & -             & -             & -             \\
T-Rex2~\cite{jiang2025t} &                                                                                        & -             & -             & 46.5          & -                         & -                         & \textbf{47.6}         & \textbf{45.4}         & 46.0                  & \textbf{49.5}          & -             & -             & -             & -             & -             & -             & -             & -             \\
Grounding DINO~\cite{liu2023grounding}    &                                                                                        & -             & -             & 48.4          & -                         & -                         & 33.0                  & 22.2                  & 30.7                  & 38.8                   & 89.2          & 91.9          & 86.0          & 81.1          & 87.4          & 74.7          & 84.2          & 84.9          \\ \hline
Shikra-7B~\cite{chen2023shikra} & \multirow{6}{*}{MLLM}                                                                  & 40.3          & 21.5          & -             & 52.8                      & 14.5                      & -                     & -                     & -                     & -                      & 87.0          & 90.6          & 80.2          & 81.6          & 87.4          & 72.1          & 82.3          & 82.2          \\
Ferret-7B~\cite{you2023ferret}         &                                                                                        & 66.3          & 33.5          & -             & 72.9                      & 25.2                      & -                     & -                     & -                     & -                      & -             & -             & -             & -             & -             & -             &               & -             \\
Groma-7B~\cite{ma2024groma}          &                                                                                        & 69.9          & 28.9          & -             & 76.3                      & 10.9                      & -                     & -                     & -                     & -                      & 89.5          & 92.1          & 86.3          & 83.9          & 88.9          & 78.1          & 86.4          & 87.0          \\
InternVL2-7B~\cite{VLM:InternVL-1.5}      &                                                                                        & 45.3          & 24.5          & -             & 51.6                      & 13.1                      & -                     & -                     & -                     & -                      & 87.1          & 91.1          & 80.7          & 79.8          & 87.9          & 71.4          & 82.7          & 82.7          \\
Qwen2-VL-7B~\cite{wang2024qwen2}       &                                                                                        & 59.3          & 43.9          & -             & 77.0                      & 34.7                      & -                     & -                     & -                     & -                      & \textbf{91.7} & 93.6          & \textbf{87.3} & 85.8          & 90.5          & \textbf{79.5} & 87.3          & 87.8          \\ \cline{1-1} \cline{3-19} 
\rowcolor{gray!15}ChatRex-7B        &                                                                                        & \textbf{73.5} & \textbf{72.8} & 48.2          & \textbf{80.3}             & \textbf{58.9}             & 42.6                  & 44.6                  & \textbf{48.4}         & 37.2                   & 91.0          & \textbf{94.1} & 87.0          & \textbf{89.8} & \textbf{91.9} & 79.3          & \textbf{89.8} & \textbf{90.0} \\ \bottomrule
\end{tabular}}
  \caption{Comparison of different models on object detection tasks on the COCO, LVIS, and RefCOCO/+/g datasets. For COCO and LVIS, we report the R@0.5 and P@0.5 metrics for MLLMs, representing recall and precision at an IoU threshold of 0.5, respectively. For RefCOCO/+/g, a prediction is considered correct if its overlap IoU with the ground truth is larger than 0.5.}
  \label{tab:detection_benchmarks}
  \end{table*}
\section{Experiments}
\label{sec:experiments}

\begin{table*}[t]
\centering
  \resizebox{\linewidth}{!}{
    \begin{tabular}{cccccccccccc}
\toprule
\multicolumn{1}{c|}{Model}             & MME    & MMB  & SEED$^I$ & MMStar & MMVet & MMMU & AI2D & OCRBench & TextVQA & POPE & Hallusion \\ \midrule
\multicolumn{1}{c|}{BLIP-2~\cite{li2023blip2}}                                 & 1293.8 & -    & 49.7 & -      & 22.4  & -    & -    & -        & -       & 85.3 & -         \\
\multicolumn{1}{c|}{InstructBLIP~\cite{VLM:InstructBLIP}}                           & 1212.8 & -    & -    & -      & -     & -    & -    & -        & -       & 78.9 & -         \\
\multicolumn{1}{c|}{Mini-Gemini-HD-8B~\cite{li2024mini}} & 1606.0 & 72.7 & 73.2 & -      & -     & 37.3 & 73.5 & 47.7     & 70.2    & -    & -         \\
\multicolumn{1}{c|}{LLaVA-HR~\cite{luo2024feast}}          & 1554.0 & -    & 64.2 & -      & 31.2  & -    & -    & -        & 67.1    & 87.6 & -         \\
\multicolumn{1}{c|}{LLaVA-NeXT-7B~\cite{VLM:LLaVA-1.6}}     & 1498.0 & 68.7 & 72.2 & 38.4   & 42.2  & 35.3 & 69.0 & 531      & 64.6    & 86.7 & 29.1      \\
\multicolumn{1}{c|}{Eagle-X5-7B~\cite{shi2024eagle}}       & 1579.0 & 68.8 & 73.5 & 41.7   & 42.6  & 36.3 & 77.2 & 574      & 71.2    & 88.8 & 37.8      \\
\multicolumn{1}{c|}{MM1.5-7B~\cite{zhang2024mm1}}          & 1514.9 & -    & 73.4 & -      & 42.2  & 41.8 & 72.2 & 635      & 76.5    & 88.6 & -         \\
\multicolumn{1}{c|}{Cambrian-8B~\cite{tong2025cambrian}}       & 1547.1 & 75.9 & 74.7 & 47.1   & 48.9  & 41.6 & 73.6 & 610      & 71.7    & 86.8 & 39.4      \\
\multicolumn{1}{c|}{LLaVA-OV-7B~\cite{li2024llava} }       & 1577.8 & 83.2 & 76.7 & 61.9   & 51.9  & 47.9 & 82.4 & 622      & 78.5    & 88.4 & 31.6      \\
\multicolumn{1}{c|}{InternVL2-8B~\cite{VLM:InternVL-1.5}}      & 1639.7 & 81.7 & 75.4 & 61.5   & 54.2  & 49.8 & 83.0 & 794      & 77.4    & 84.2    & 45.0      \\
\multicolumn{1}{c|}{Qwen2-VL-7B~\cite{wang2024qwen2} }       & 1639.2 & 83.0 & 76.0 & 60.7   & 62.0  & 54.1 & 83.0 & 845      & 84.3    & 88.4 & 50.6      \\
\rowcolor{gray!15}\multicolumn{1}{c|}{ChatRex-7B}        & 1544.0 & 81.1 & 74.4 & 57.5   & 41.5  & 46.7 & 79.1 & 626      & 69.1    & 87.6 & 39.1      \\ \bottomrule
\end{tabular}}
  \caption{Comparison of different models on multimodal benchmarks.}
  \label{tab:multimodal_benchmarks}
  \end{table*}

\subsection{Perception Capability Evaluation}
\textbf{Evaluation Metrics}. Mean Average Precision (mAP)~\cite{Datasets:MSCOCO} is a common metric for object detection, which measures the area under the precision-recall curve, reflecting both the precision and recall of the model. However, for MLLMs that predict coordinates as vocabulary tokens, computing AP can be challenging due to the lack of confidence scores for each predicted box. Therefore, we directly report recall and precision metrics instead. We provide all ground truth categories for the current test image and prompt the model to generate the corresponding coordinate boxes. The details of prompts used for each model are included in the Appendix. For ChatRex, we use fine-grained proposal boxes from UPN and their corresponding confidence scores as input, enabling us to compute precision, recall, and mAP.

\textbf{Common Object Detection}. As shown in Tab. \ref{tab:detection_benchmarks}, ChatRex achieves a 48.2 mAP on the COCO dataset, which is comparable to conventional object detectors like DINO~\cite{zhang2022dino}, indicating that ChatRex possesses strong perception capabilities. In contrast, other MLLMs generally exhibit low recall rates. This discrepancy arises from the multi-object nature of COCO, where each image contains multiple categories with numerous instances. The low recall rate implies that current MLLMs face significant challenges in detecting multiple objects, which is a common requirement in real-world scenarios. Furthermore, we identified specific issues with general MLLMs such as InternVL2 and Qwen2-VL, which have a tendency to repeatedly generate the same coordinates until reaching the model's maximum output length. A detailed analysis of these problems is provided in the Appendix, highlighting areas for potential improvement in future work.

\textbf{Long-tailed Object Detection}. We further evaluate ChatRex on the more challenging LVIS~\cite{gupta2019lvis} dataset, which encompasses 1,203 object categories. ChatRex achieved a 42.6 mAP, surpassing open-set detection models like Grounding DINO~\cite{liu2023grounding} and GLIP~\cite{li2022grounded} and is on par with T-Rex2~\cite{jiang2025t}. We attribute this performance to the strong semantic understanding capabilities of the LLM. Within the ChatRex model structure, the LLM primarily functions to classify bounding boxes generated by the proposal model. By aligning visual features with the textual space through comprehensive training and data optimization, the LLM is able to accurately classify a broad spectrum of categories, thereby demonstrating its robustness in handling complex, long-tailed object detection scenarios.

\textbf{Referring Object Detection}. Referring object detection involves identifying an object based on a given description. We evaluate ChatRex on the RefCOCO, RefCOCO+, and RefCOCOg benchmarks, which predominantly focus on single-object detection, where each expression generally corresponds to a single object. As shown in Tab. \ref{tab:detection_benchmarks}, ChatRex possesses strong referring capabilities, which are crucial for tackling complex perception tasks.

\subsection{Understanding Capability Evaluation}
\textbf{General Multimodal Benchmarks}. We evaluate ChatRex on various academic multimodal benchmarks including MME~\cite{fu2024mmecomprehensiveevaluationbenchmark}, MMBench~\cite{Datasets:MMBench}, SEED$^{I}$~\cite{li2023seed}, MMstar~\cite{chen2024we}, MM-Vet~\cite{yu2023mmvet}, MMMU~\cite{yue2023mmmu}, AI2D~\cite{kembhavi2016diagram}, OCRBench~\cite{liu2024ocrbench}, TextVQA~\cite{singh2019towards}, POPE~\cite{Datasets:POPE}, and HallusionBench~\cite{guan2023hallusionbench}. As shown in Table \ref{tab:multimodal_benchmarks}, ChatRex demonstrates strong multimodal capabilities, though there remains a performance gap compared to state-of-the-art models like Qwen2-VL and InternVL2. We believe this performance gap can be narrowed with larger-scale data and improved visual representations, such as the AnyRes strategy~\cite{li2024llava}. Furthermore, ChatRex’s enhanced perception capabilities enable it to provide object grounding information during conversations, extending its applicability to a broader range of real-world scenarios.

\textbf{Region Caption Benchmarks}. In addition to image-level understanding, ChatRex demonstrates strong region-level understanding capabilities. Following Osprey~\cite{yuan2024osprey}, we evaluate the referring object classification task on the LVIS~\cite{gupta2019lvis} and PACO~\cite{ramanathan2023paco} datasets. In this task, given the object index, the model is prompted to output the category name of the specified region. The evaluation metrics include Semantic Similarity (SS) and Semantic Intersection over Union (S-IOU)~\cite{rezatofighi2019generalized}. As shown in Tab. \ref{tab:roc}, ChatRex achieves state-of-the-art results, highlighting its robust region classification capabilities.

\begin{table}[]
  \centering
  \resizebox{0.8\columnwidth}{!}{%
   \begin{tabular}{c|cc|cc}
\toprule
\multirow{2}{*}{Method} & \multicolumn{2}{l|}{LVIS} & \multicolumn{2}{l}{PACO} \\
                        & SS          & S-IoU       & SS         & S-IoU       \\ \midrule
LLaVA-1.5~\cite{VLM:LLaVA}               & 49.0        & 19.8        & 42.2       & 14.6        \\
Kosmos-2~\cite{peng2023kosmos2}               & 39.0        & 8.7         & 32.1       & 4.8         \\
Shikra-7B~\cite{chen2023shikra}               & 49.7        & 19.8        & 43.6       & 11.4        \\
GPT4RoI-7B~\cite{VLM:GPT4ROI}              & 51.3        & 12.0        & 48.0       & 12.1        \\
Ferret-7B~\cite{you2023ferret}               & 63.8        & 36.6        & 58.7       & 26.0        \\
Osprey-7B~\cite{yuan2024osprey}               & 65.2        & 38.2        & 73.1       & 52.7        \\
VisionLLM v2-7B~\cite{wu2024visionllm}         & 68.9        & 46.3        & 67.7       & 44.0        \\
SPHINX-V-7B~\cite{lin2023sphinx}             & 87.1        & 62.9        & 80.0       & 55.0        \\
\rowcolor{gray!15} ChatRex-7B (Ours)       & \textbf{89.8}        & \textbf{82.6}        & \textbf{91.4}       & \textbf{85.1}        \\ \bottomrule
\end{tabular}}
\caption{Comparison on referring object classification task.}
\label{tab:roc}
\end{table}

\begin{table}[]
  \centering
  \resizebox{1.0\columnwidth}{!}{%
   \begin{tabular}{cccccccc}
\hline
\multicolumn{1}{c|}{\multirow{2}{*}{\begin{tabular}[c]{@{}c@{}}Training\\ With Stage-2\end{tabular}}} & \multicolumn{7}{c}{Understanding}                                                                                                                                                                                                                                                                                                                                                     \\ \cline{2-8} 
\multicolumn{1}{c|}{}                                                                                & MME                                                & MMB                                                & MMStar                                             & SEED$^I$                                                 & MMMU                                                 & POPE                                                & Hallusion                                          \\ \hline
\multicolumn{1}{c|}{No}                                                                              & 1388.8                                             & 76.8                                               & 44.5                                               & 72.9                                                 & 46.0                                                 & 80.3                                                & 34.0                                               \\
\multicolumn{1}{c|}{Yes}                                                                             & 1439.1                                             & 76.7                                               & 46.2                                               & 73.0                                                 & 52.0                                                 & 86.0                                                & 37.6                                               \\ \hline
                                                                                                     & \multicolumn{7}{c}{Perception}                                                                                                                                                                                                                                                                                                                                                        \\ \hline
\multicolumn{1}{c|}{\begin{tabular}[c]{@{}c@{}}Training\\ With Stage-2\end{tabular}}                  & \begin{tabular}[c]{@{}c@{}}COOC\\ mAP\end{tabular} & \begin{tabular}[c]{@{}c@{}}LVIS\\ mAP\end{tabular} & \begin{tabular}[c]{@{}c@{}}Ref+\\ val\end{tabular} & \begin{tabular}[c]{@{}c@{}}Ref+\\ testA\end{tabular} & \begin{tabular}[c]{@{}c@{}}Ref+\\ testB\end{tabular} & \begin{tabular}[c]{@{}c@{}}Refg\\ test\end{tabular} & \begin{tabular}[c]{@{}c@{}}Refg\\ val\end{tabular} \\ \hline
\multicolumn{1}{c|}{No}                                                                              & 47.8                                               & 42.3                                               & 85.9                                               & 90.9                                                 & 79.7                                                 & 89.4                                                & 88.8                                               \\
\multicolumn{1}{c|}{Yes}                                                                             & 48.7                                               & 43.4                                               & 88.6                                               & 92.2                                                 & 82.2                                                 & 90.6                                                & 90.3                                               \\ \hline
\end{tabular}}
\caption{Ablation of the impact of Stage-2 training on the perception and understanding capabilities of ChatRex. We remove the LLaVA-Onevision data in the final stage for quick validation.}
\label{tab:ablation_stage2}
\end{table}

% \begin{table}[]
%   \centering
%   \resizebox{1.0\columnwidth}{!}{%
%    \begin{tabular}{c|cccc}
% \toprule
% \begin{tabular}[c]{@{}c@{}}With Separated\\ Grained Training\end{tabular} & \begin{tabular}[c]{@{}c@{}}COCO\\ R@0.3\end{tabular} & \begin{tabular}[c]{@{}c@{}}COCO\\ mAP\end{tabular} & \begin{tabular}[c]{@{}c@{}}LVIS\\ R@0.3\end{tabular} & \begin{tabular}[c]{@{}c@{}}LVIS\\ mAP\end{tabular} \\ \midrule
% No                                                                        & 87.3                                                 & 47.4                                               & 67.9                                                 & 35.3                                               \\
% Yes                                                                       & 91.2                                                 & 48.5                                               & 77.1                                                 & 43.1                                               \\ \bottomrule
% \end{tabular}}
% \caption{Ablation on the granularity-based training design of the universal proposal network. R@0.3 and P@0.3 represents recall and precision at score threshold at 0.3.}
% \label{tab:ab_proposal}
% \end{table}

\begin{table}[]
  \centering
  \resizebox{1.0\columnwidth}{!}{%
   \begin{tabular}{c|ccccccc}
\hline
Eval Stage & \begin{tabular}[c]{@{}c@{}}COCO\\ mAP\end{tabular} & \begin{tabular}[c]{@{}c@{}}LVIS\\ mAP\end{tabular} & \begin{tabular}[c]{@{}c@{}}Ref+\\ val\end{tabular} & \begin{tabular}[c]{@{}c@{}}Ref+\\ testA\end{tabular} & \begin{tabular}[c]{@{}c@{}}Ref+\\ testB\end{tabular} & \begin{tabular}[c]{@{}c@{}}Refg\\ test\end{tabular} & \begin{tabular}[c]{@{}c@{}}Refg\\ val\end{tabular} \\ \hline
Stage-2     & 48.7                                               & 42.2                                               & 86.5                                               & 91.9                                                 & 80.9                                                 & 89.6                                                & 89.0                                               \\
Stage-3     & 48.2                                               & 42.6                                               & 89.8                                               & 91.9                                                 & 79.3                                                 & 89.8                                                & 90.0                                               \\ \hline
\end{tabular}}
\caption{Ablation of the perception capabilities of ChatRex at different stage checkpoints.}
\label{tab:ablation_stage2_stage3}

\end{table}

\begin{table}[]
  \centering
  \resizebox{1.0\columnwidth}{!}{%
   \begin{tabular}{cccccccc}
\hline
\multicolumn{1}{c|}{\multirow{2}{*}{Method}} & \multicolumn{7}{c}{Understanding}                                                                                                                                                                                                                                                                                                                                                     \\ \cline{2-8} 
\multicolumn{1}{c|}{}                        & MME                                                & MMB                                                & MMStar                                             & SEED$^I$                                                 & MMMU                                                 & POPE                                                & Hallusion                                          \\ \hline
\multicolumn{1}{c|}{Baseline}                & 1439.1                                             & 76.7                                               & 46.2                                               & 73.0                                                 & 52.0                                                 & 86.0                                                & 37.6                                               \\
\multicolumn{1}{c|}{w/o ConvNeXt}            & 1341.9                                             & 70.4                                               & 42.7                                               & 65.4                                                 & 43.3                                                 & 83.8                                                & 24.4                                               \\ \hline
                                             & \multicolumn{7}{c}{Perception}                                                                                                                                                                                                                                                                                                                                                        \\ \hline
\multicolumn{1}{c|}{Method}                  & \begin{tabular}[c]{@{}c@{}}COOC\\ mAP\end{tabular} & \begin{tabular}[c]{@{}c@{}}LVIS\\ mAP\end{tabular} & \begin{tabular}[c]{@{}c@{}}Ref+\\ val\end{tabular} & \begin{tabular}[c]{@{}c@{}}Ref+\\ testA\end{tabular} & \begin{tabular}[c]{@{}c@{}}Ref+\\ testB\end{tabular} & \begin{tabular}[c]{@{}c@{}}Refg\\ test\end{tabular} & \begin{tabular}[c]{@{}c@{}}Refg\\ val\end{tabular} \\ \hline
\multicolumn{1}{c|}{Baseline}                & 48.7                                               & 43.4                                               & 88.6                                               & 92.2                                                 & 82.2                                                 & 90.6                                                & 90.3                                               \\
\multicolumn{1}{c|}{w/o ConNeXt}             &            26.0                                        &                           22.0                         &                                     66.8               & 73.2                                                 & 61.4                                                 & 73.1                                                & 72.5                                               \\ \hline
\end{tabular}}
\caption{Ablation of the dual vision encoder designing in ChatRex. We evaluate the impact of removing the ConvNeXt high-resolution encoder on perception and understanding tasks. We remove the LLaVA-Onevision data in Stage-3 for quick validation.}
\label{tab:ab_highresolution}
\vspace{-1.5em}
\end{table}

\subsection{Ablation Experiments}

\textbf{Mutual Benefits of Perception and Understanding}.
We conduct ablation experiments to analyze the mutual influence of perception and understanding in ChatRex. As shown in Tab. \ref{tab:ablation_stage2}, incorporating perception training in Stage-2 enhances the model’s multimodal understanding, demonstrating that stronger perception contributes to improved understanding. Additionally, training with Stage-2 also leads to a performance boost in perception, highlighting the effectiveness of the Rexverse-2M dataset in strengthening perception capabilities.

To further examine this interaction, we compare the performance of the perception between Stage-2 and Stage-3, as shown in Tab. \ref{tab:ablation_stage2_stage3}. The results indicate that the perception performance of ChatRex improves with Stage-3 training, suggesting that multimodal data not only enhances understanding but also refines perception. This reinforces the interdependence of perception and understanding, demonstrating that their integration leads to mutual enhancement and a more robust multimodal model.

\noindent\textbf{Ablation on ChatRex Architecture}.
In ChatRex, we adopt a dual vision encoder design, where object features are extracted from the high-resolution encoder to serve as object tokens. To assess the effectiveness of this design, we conduct an ablation study by removing the high-resolution encoder and instead extracting object features from the low-resolution CLIP encoder. As shown in Tab. \ref{tab:ab_highresolution}, the results indicate that the removal of the high-resolution encoder leads to a decline in both perception and understanding performance, with a more significant drop observed in perception. We attribute this to the fact that perception tasks heavily rely on higher input image resolutions to capture fine-grained details, which aligns with established findings in object detection community, where increasing image resolution generally improves detection accuracy and overall performance.
% \noindent\textbf{Ablation on Universal Proposal Network}.
% In ChatRex, we leverage a Universal Proposal Network (UPN) to supply box inputs to the LLM. We conduct an ablation study to validate the effectiveness of the granularity-based training methods. Specifically, we compare results with and without the use of distinct granularity prompts. In the absence of dual granularity, all datasets are treated as a single class during training. We evaluate the recall rates on the COCO and LVIS datasets at a fixed confidence threshold of 0.3, as well as their final mAP. As shown in Tab. \ref{tab:ab_proposal}, the granularity-based approach achieves a higher recall rate, indicating that proposal models trained with separate granularity prompts can detect a larger number of objects at a given confidence level, ultimately enhancing the overall detection performance.

%% file: sec/6_Conclusion.tex
\section{Conclusions}
\label{sec:conclusions} 
In this work, we reveal the perception drawback in existing MLLMs due to their conflict modeling between perception and understanding, and the lack of data that effectively balance these two aspects. To address these issues, we introduce ChatRex, a model with a decoupled architecture, along with Rexverse-2M, a multi-granularity dataset designed to balance these two aspects. ChatRex demonstrates strong perception abilities while also excelling in multimodal understanding and dialog capabilities. The synergy between perception and understanding allows ChatRex to be highly interactive by grounding mentioned objects within visual contexts during dialogue. We believe that perception and understanding are both critical for MLLMs, as their integration can significantly enhance model capabilities and unlock a wide range of novel applications.

%% file: sec/X_suppl.tex
\clearpage

\setcounter{page}{1}
\maketitlesupplementary

\begin{figure*}[t]\centering
\includegraphics[width=0.91\linewidth]{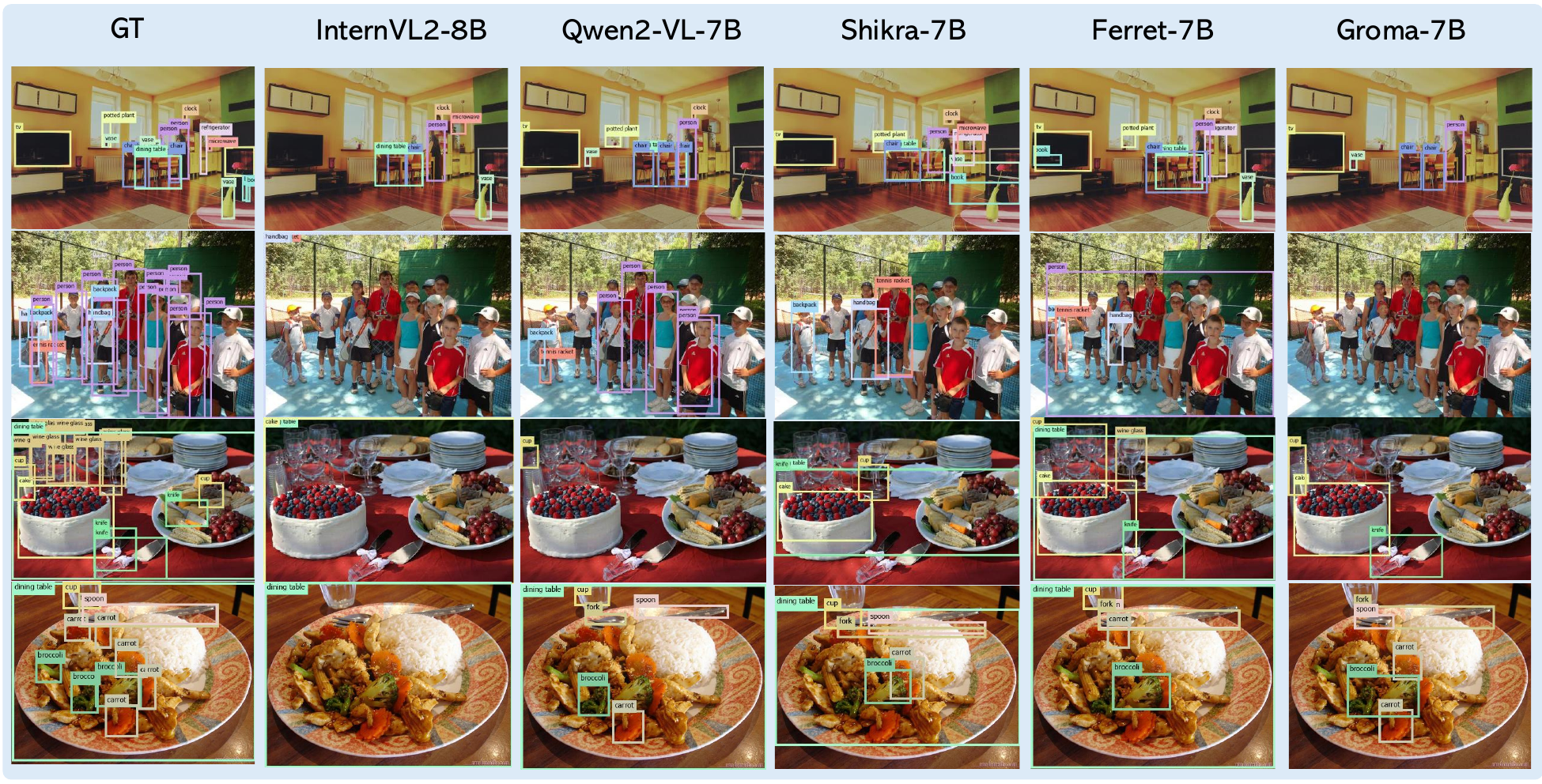}\vspace{-1mm}
\caption{Visualized prediction results on the COCO dataset, from general-purpose MLLM including Qwen2-VL~\cite{wang2024qwen2}, InternVL2~\cite{VLM:InternVL-1.5}, and detection-focused MLLMs including Ferret~\cite{you2023ferret}, Shikra~\cite{chen2023shikra}, and Groma~\cite{ma2024groma} These models generally suffer from a low recall rate in multi-object scenes.}
\label{fig:mllms_on_coco}
\vspace{-1mm}
\end{figure*}

\renewcommand{\thesection}{\Alph{section}}
\setcounter{section}{0}

\section{MLLM Evaluation Details on Detection Datasets}
\label{sec:rationale}
In this section, we explain our methodology for evaluating MLLMs on the object detection task, including the design of model-specific prompts, the visualization of output results, and a comprehensive analysis of the challenges and limitations encountered throughout the evaluation process.

\subsection{Prompt for Each MLLM}
For each MLLM evaluated, we utilized either the prompts used in the original paper or manually crafted optimized prompts to maximize performance. The specific prompts used in our evaluation are detailed in Tab. \ref{tab:mllm_prompts},

\begin{table*}[t]
\centering
  \resizebox{\linewidth}{!}{
    \begin{tabular}{c|ll}
\toprule
Method                                                                 & Prompt                                                                                                                                                                                                                                                                                                                                                                                                                                                                                                                                                                                                                                                                                                                                                                                                                                                                                                                                                                                                                                                                                                                                                                    &  \\ \midrule
\begin{tabular}[c]{@{}c@{}}Qwen2-VL-7B \& \\ InternVL2-8B\end{tabular} & \begin{tabular}[c]{@{}l@{}}In this picture, you are required to finish object detection for every instance of the category we provide. To complete the above mission,\\ you need to provide me with the answers in the format of a Python list of dictionaries by the category provided above. Attention: No \\ other category shall appear in the detection object attributes, except for the genre we offer.  Bounding box format: {[}108(xmin), 210(ymin),\\ 810(xmax), 640(ymax){]}, where xmin, ymin, xmax and ymax must be positive integers. If there is no object in the picture, please provide\\ an empty list. Here is an example which you must follow in your responses. Example: If the question is as below: Category: {[}'person', 'car'{]}.\\ If there is an object of the category, The Answer should be: {[}\{\{"class": "person", "rect": {[}0, 614, 220, 771{]}\}\}, \{\{"class": "person", "rect":\\ {[}638, 468, 784, 941{]}\}\}, \{\{"class": "car", "rect": {[}110, 100, 500, 300{]}\}\}{]}. Else if no object of the category in the picture, the Answer should\\ be:{[}{]}. Here is the question you shall answer: Category: \{\}\end{tabular} &  \\ \hline
Ferret-7B                                                               & What is the location of all instances of categories \{\} in the image? Please answer me respectively.                                                                                                                                                                                                                                                                                                                                                                                                                                                                                                                                                                                                                                                                                                                                                                                                                                                                                                                                                                                                                                                                     &  \\ \midrule
Shikra-7B                                                              & Help me locate \{\} in \textless{}image\textgreater and give its coordinates, please.                                                                                                                                                                                                                                                                                                                                                                                                                                                                                                                                                                                                                                                                                                                                                                                                                                                                                                                                                                                                                                                                                     &  \\ \midrule
Groma-7B                                                               & \begin{tabular}[c]{@{}l@{}}{[}grounding{]} There are categories you need to describe with positions, only including \textless{}p\textgreater{}\{\}\textless{}p\textgreater{}. Give me a short description of the image\\ and include the coordinates {[}{[}x0,y0,x1,y1{]}{]} for each instance of categories.\end{tabular}                                                                                                                                                                                                                                                                                                                                                                                                                                                                                                                                                                                                                                                                                                                                                                                                                                                &  \\ \bottomrule
\end{tabular}}
  \caption{Prompt used by each MLLM for object detection.}
  \label{tab:mllm_prompts}
  \end{table*}

\begin{table*}[h]
\centering
  \resizebox{\linewidth}{!}{
    \begin{tabular}{c|c}
\toprule
Task                                  & Prompt Templates                                                                                                                           \\ \midrule
Grounding \& Detection \& Referring   & Please detect {[}OBJ{]} in this image. Answer the question with object indexes.                                                            \\
Brief Image Caption with Grounding    & Please briefly describe this image and detect all the mentioned objects. Answer with grounded object indexes.                              \\
Detailed Image Caption with Grounding & Please provide a detailed description of the image and detect all the mentioned objects. Answer the question with grounded object indexes. \\
Region Caption in Category Name       & What is the category name of {[}OBJ{]}? Answer the question with its category name in free format.                                         \\
Region Caption in Short Phrase        & Can you provide me with a short phrase description of {[}OBJ{]}? Answer the question with short phrases.                                   \\
Region Caption Briefly                & Can you provide me with a brief description of {[}OBJ{]}? Answer the question with brief description.                                      \\
Region Caption in One Sentence        & Can you provide a one sentence description of {[}OBJ{]} in the image? Answer the question with one sentence description.                   \\
Grounded Counting                     & How many {[}OBJ{]} are there in this image? Answer the question with the number of objects and locate them with object indexes.            \\
Grounded Conversation                 & Answer the question in grounded format. Question:                                                                                          \\ \bottomrule
\end{tabular}}
  \caption{Example prompt for different perception tasks of ChatRex.}
  \label{tab:special_prompts}
  \end{table*}

\begin{table*}[h]
\centering
  \resizebox{\linewidth}{!}{
    \begin{tabular}{c|c|cl}
\toprule
Task                                                                            & Model              & Prompt                                                                                                                                                                                                                                                                                                                                                                                                                                                                                                                                                                                                                                                                                                             &  \\ \midrule
Image Description                                                               & InternVL2-8B       & Please provide a one-sentence description for this image.                                                                                                                                                                                                                                                                                                                                                                                                                                                                                                                                                                                                                                                          &  \\ \midrule
\begin{tabular}[c]{@{}c@{}}Phrase-Conditioned \\ Region Descrption\end{tabular} & InternVL2-8B       & \begin{tabular}[c]{@{}c@{}}I will provide you with a short phrase description of an object and its image. You need to rewrite this short\\ phrase description to a one sentence description by adding more details about this object based on the image.\\ The rewritten description can only focus on this object according to the original description and should also\\  be a one-sentence description. The original short phrase description is:\end{tabular}                                                                                                                                                                                                                                                  &  \\ \midrule
\begin{tabular}[c]{@{}c@{}}Region Filtering \&\\ Rewriting\end{tabular}         & LLAMA3-8B-Instruct & \begin{tabular}[c]{@{}c@{}}I will provide you with a one-sentence description of an object, and the category name of that object. Based\\ on these two pieces of information, write a referring description of the object. This description should capture\\ the most important and distinguishing features of the object, and should not describe anything that doesn't\\ exist in the description I've provided. Note that the referring object should be the category name provided.\\ The rewritten referring description should be more than 5 words but less than 10 words. The referring\\ description should be as short and concise as possible, without commas. Directly output the answer.\end{tabular} &  \\ \bottomrule
\end{tabular}}
  \caption{Prompt for each module where MLLM or LLM is used.}
  \label{tab:engine_prompt}
  \end{table*}

\subsection{More Analysis on the Detection Results}

The visualization results on the COCO dataset are presented in Fig. \ref{fig:mllms_on_coco}. The qualitative analysis indicates that most evaluated MLLMs suffer from a low recall rate, underscoring significant limitations in their object perception capabilities. Additionally, specific deficiencies are observed in certain models. For example, Shikra~\cite{chen2023shikra} exhibits a notable issue with coordinate offset, where the predicted bounding boxes fail to accurately enclose the target objects. This misalignment is primarily attributed to quantization errors inherent in the model. Moreover, during the evaluation phase, most models, excluding general-purpose MLLMs such as InternVL2~\cite{VLM:InternVL-1.5} and Qwen2-VL~\cite{wang2024qwen2}, frequently struggle to follow task-specific instructions. This observation highlights persistent challenges in the instruction-following capability within these models.

We also identified a distinct issue with general MLLMs during the evaluation process: a pattern of repeated output coordinates in their predictions. This repetition is not random but follows a systematic pattern. For example, in the case of Qwen2-VL-7B, the model consistently generates bounding box coordinates where each subsequent bounding box is derived by incrementally adjusting the preceding one. For example, the \texttt{xmin} and \texttt{xmax} values of the bounding boxes are increased by a fixed value, such as 24, as illustrated in the below block. We argue this behavior stems from the auto-regressive coordinate prediction mechanism. The reliance on previously predicted coordinates introduces a probabilistic bias, causing subsequent predictions to deviate systematically as regular modifications of earlier outputs. This limitation underscores the model's inability to independently and accurately generate coordinates without cumulative error propagation.

\begin{tcolorbox}[colback=black!5!white,colframe=black!75!black,title=Example of repeated coordinate prediction in Qwen2-VL-7B.]

\textbf{Response}:
[
\{class: car, rect: [234, 186, 370, 283]\}, \\
\{class: car, rect: [568, 214, 622, 283]\}, \\
\{class: car, rect: [743, 186, 822, 300]\}, \\
\{class: car, rect: [110, 199, 128, 240]\}, \\
\{class: car, rect: [134, 200, 152, 240]\}, \\ 
\{class: car, rect: [158, 200, 176, 240]\}, \\ 
\{class: car, rect: [182, 200, 200, 240]\}, \\
\{class: car, rect: [206, 200, 224, 240]\} \\
]... Repeat until exceeds the max output length.

\end{tcolorbox}

\section{Model Details}
\subsection{Implementation Details of UPN}

In alignment with the methodology of T-Rex2~\cite{jiang2025t}, we employ the Swin Transformer Large model~\cite{liu2021swin}, pre-trained on ImageNet~\cite{deng2009imagenet}, as the vision backbone. During the Hungarian matching process, the optimization incorporates three types of losses: classification loss, box L1 loss, and generalized intersection over union (GIOU) loss~\cite{rezatofighi2019generalized}, with respective weights of 2.0, 5.0, and 2.0. For the overall loss computation, we similarly utilize classification loss, box L1 loss, and GIOU loss, adjusting the corresponding weights to 1.0, 5.0, and 2.0. Consistent with the training strategy of DINO~\cite{zhang2022dino}, we adopt contrastive denoising training (CDN) to enhance training stability and accelerate convergence. The pre-trained weights of T-Rex2-L are used for initialization, followed by full-parameter optimization on the universal proposal task.

% \subsection{Visualization Results of UPN}
% In Fig. \ref{fig:vis_proposal}, we present visualization results highlighting the performance of our Universal Proposal Network (UPN) in generating fine-grained universal proposals. The UPN exhibits a remarkable ability to detect objects in a wide variety of scenes, underscoring its adaptability and robustness. This capability serves as a crucial foundation for enabling the LLM to identify and reference any object by leveraging the corresponding box indices produced by the UPN.

% \begin{figure*}[h]\centering
% \includegraphics[width=0.91\linewidth]{images/vis_universal_proposal.pdf}\vspace{-1mm}
% \caption{Visualization results of the proposed fine-grained universal proposal.}
% \label{fig:vis_proposal}
% \vspace{-1mm}
% \end{figure*}

\subsection{Implementation Details for ChatRex}
We utilize the CLIP pre-trained ViT-Large-14-336 model as the low-resolution visual encoder and the LAION pre-trained ConvNext-Large-320 model as the high-resolution visual encoder. The input resolution is set to 336x336 for the low-resolution encoder and 768x768 for the high-resolution encoder. During the pretraining stage, we employ a batch size of 32 per device, resulting in an aggregate batch size of 256 across all devices. For the instruction-tuning stage, the batch size is reduced to 16 per device, with a total batch size of 128. The learning rate is initialized at 1e-3 for the pre-training stage and adjusted to 2e-5 during the instruction-tuning stage.

For perception and region-based question-answering tasks, we designed tailored prompts to effectively guide and instruct the models. Examples of these customized prompts are provided in Tab. \ref{tab:special_prompts}.

\section{Details for the Rexverse-2M Data Engine}

\subsection{Visualization of Rexverse-2M dataset}
\begin{figure*}[t]\centering
\includegraphics[width=0.91\linewidth]{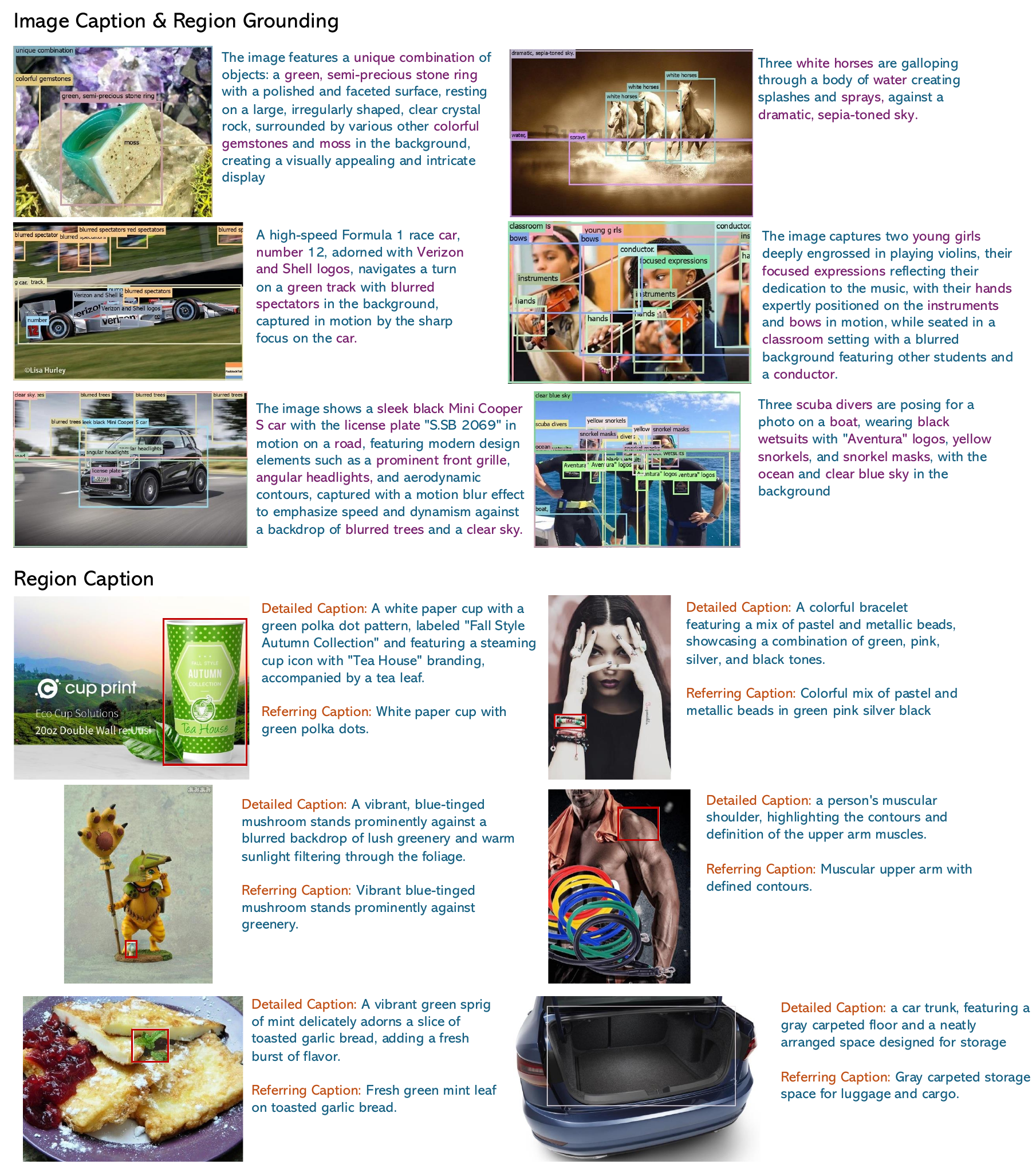}\vspace{-1mm}
\caption{Visualization of the Rexverse-2M dataset.}
\label{fig:rexverse2m}
\vspace{-1mm}
\end{figure*}

We visualize a portion of the Rexverse-2M dataset in Fig. \ref{fig:rexverse2m}, including image-level annotations and region-level annotations.

\subsection{Prompt for Different Modules}
In the Rexverse-2M data engine, we leverage both state-of-the-art MLLMs and Large Language Models LLMs to construct the dataset. The prompts employed in each module are detailed in Tab. \ref{tab:engine_prompt}.
\subsection{Effectiveness of Phrase-Conditioned Region Description}

\begin{figure*}[t]\centering
\includegraphics[width=0.91\linewidth]{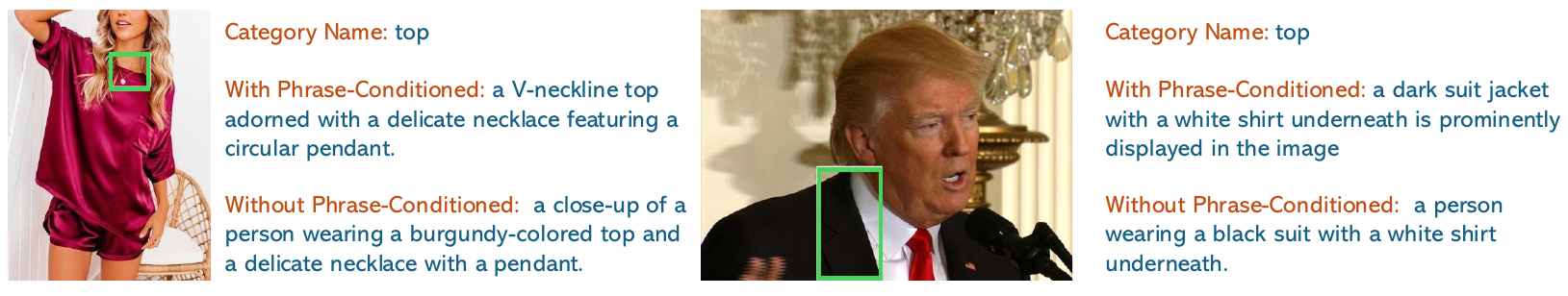}\vspace{-1mm}
\caption{Visualization of the proposed phrase-conditioned region description method.}
\label{fig:phraes_condtioned}
\vspace{-1mm}
\end{figure*}

In the data engine, to generate region-level descriptions, we adopt a phrase-conditioned region captioning method. This approach utilizes a short phrase or category name, specified for each region during the grounding phase, as an additional prompt to guide a multimodal large language model (MLLM) in producing captions. This method differs from directly inputting regional images into the MLLM for captioning. As illustrated in Figure \ref{fig:phraes_condtioned}, the direct image captioning approach often suffers from hallucinations, particularly when regions contain distracting objects or are too small to be reliably recognized. In contrast, the phrase-conditioned method mitigates these issues by incorporating contextual input, leading to more accurate captions.

\section{ChatRex Capabilities and Qualitative Examples}
In this section, we present the visualization results to demonstrate the capabilities of ChatRex.
\subsection{Common Object Detection}

\begin{figure*}[t]\centering
\includegraphics[width=0.91\linewidth]{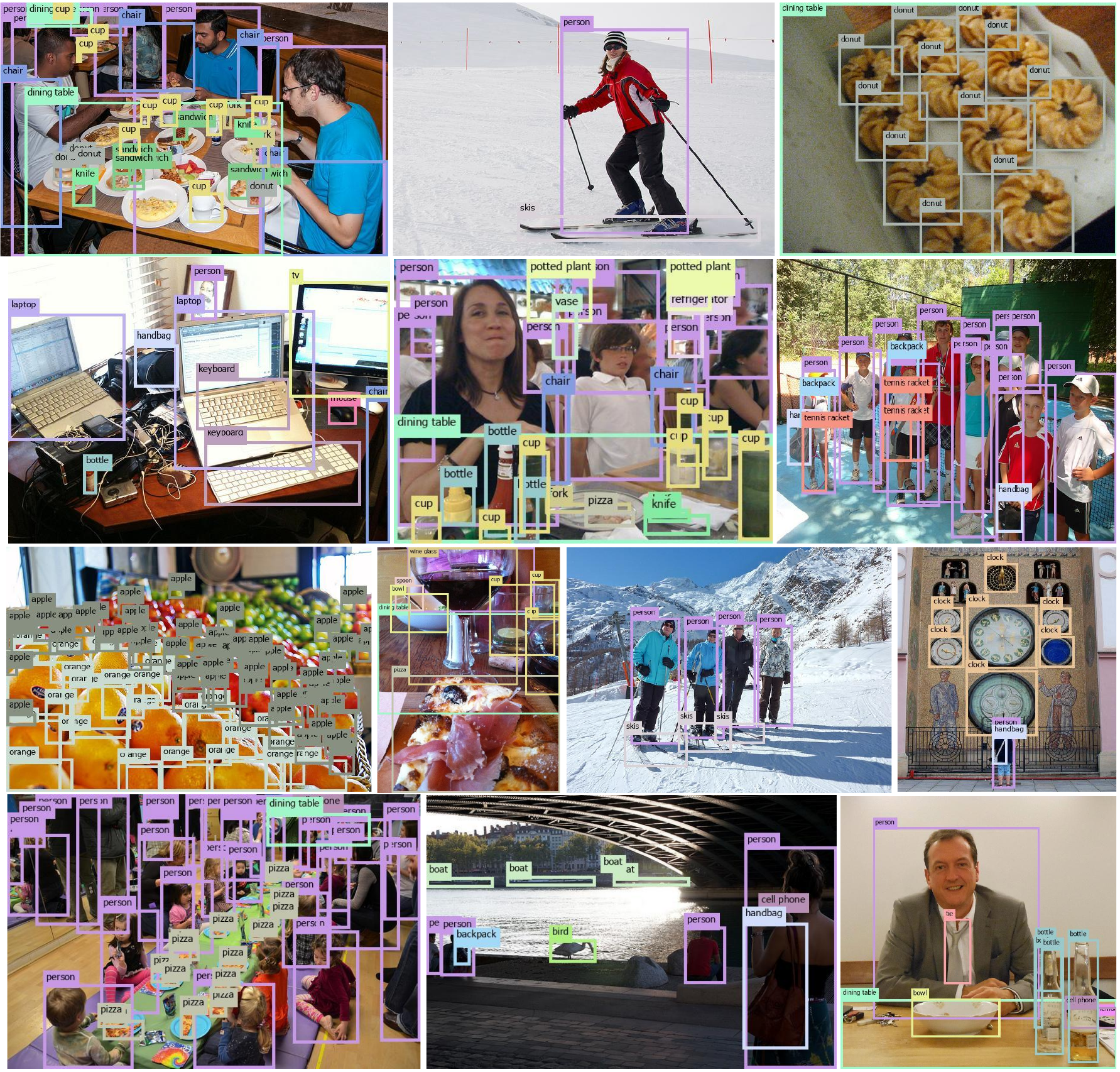}\vspace{-1mm}
\caption{Visualization on the Common Object Detection Task.}
\label{fig:vis_common_od}
\vspace{-1mm}
\end{figure*}

We show the results on the common object detection task in Fig. \ref{fig:vis_common_od}.

\subsection{Long-tailed Object Detection}

\begin{figure*}[t]\centering
\includegraphics[width=0.91\linewidth]{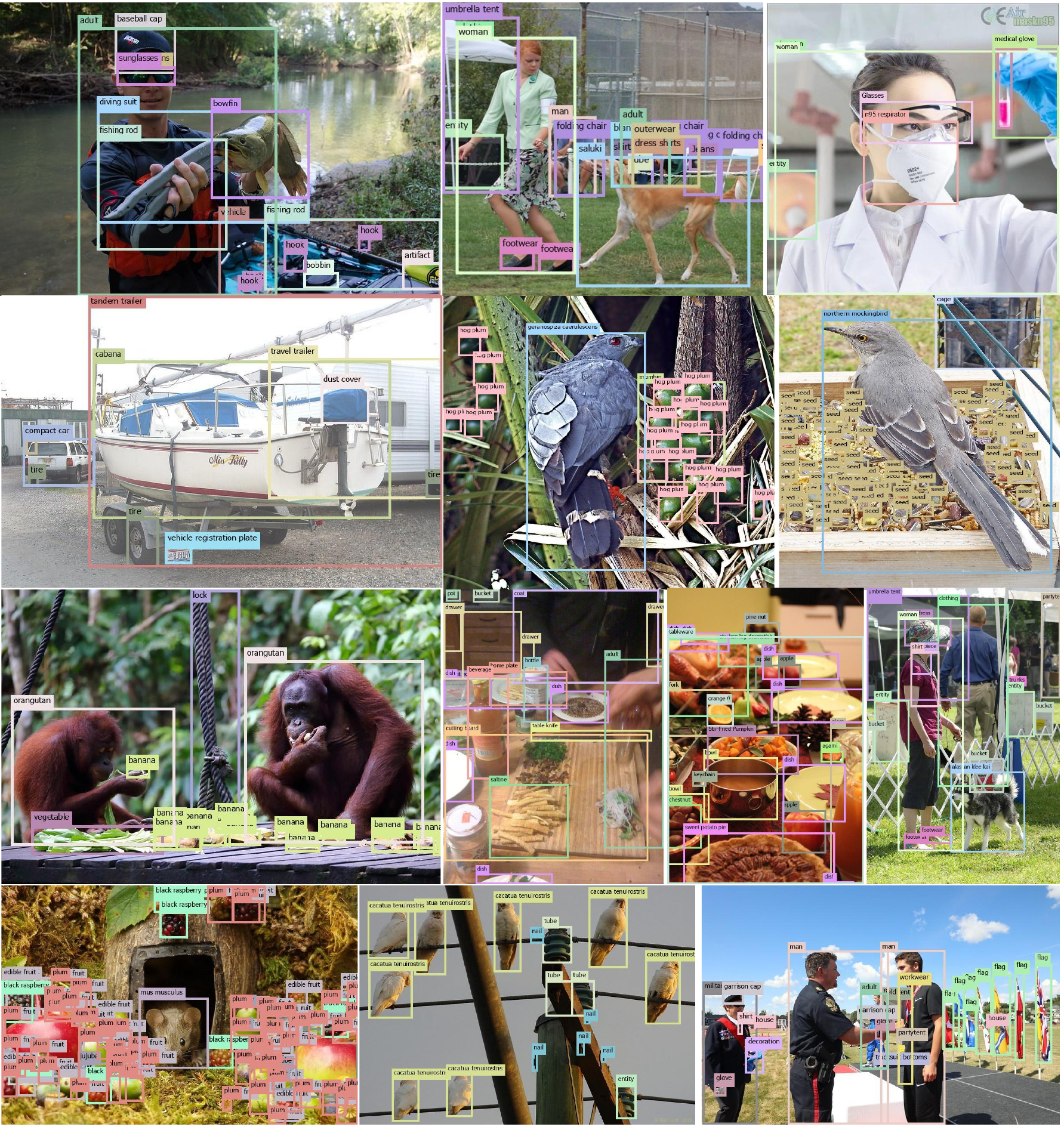}\vspace{-1mm}
\caption{Visualization on the Long-tailed Object Detection Task.}
\label{fig:vis_long_tail_detection}
\vspace{-1mm}
\end{figure*}

We show the results on the long-tailed object detection task in Fig. \ref{fig:vis_long_tail_detection}.

\subsection{Short-Phrase Object Detection}

\begin{figure*}[t]\centering
\includegraphics[width=0.91\linewidth]{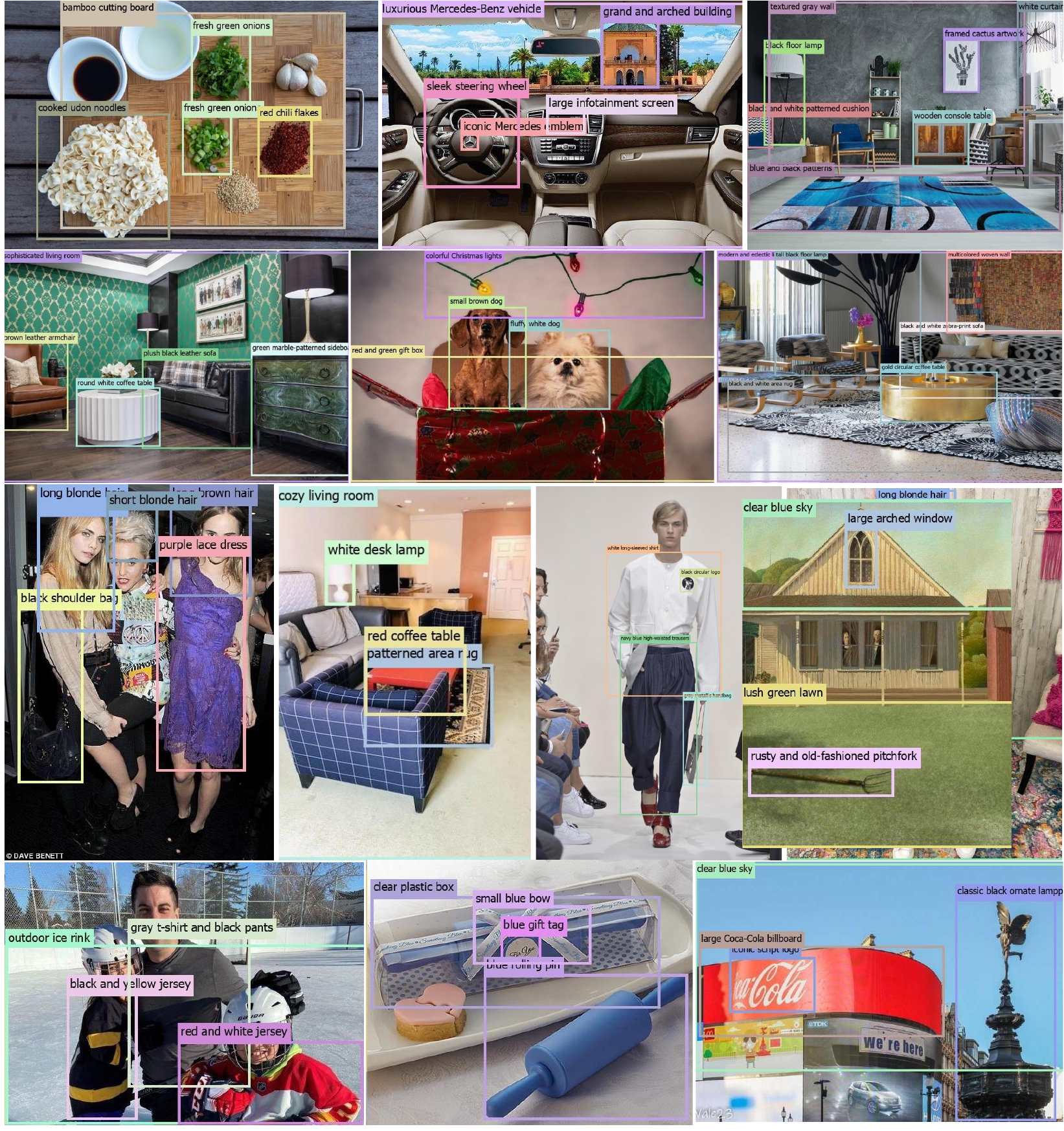}\vspace{-1mm}
\caption{Visualization on the Short-Phrase Object Detection Task.}
\label{fig:vis_short_phrase_det}
\vspace{-1mm}
\end{figure*}

We show the results on the short-phrase object detection task in Fig. \ref{fig:vis_short_phrase_det}.

\subsection{Referring Object Detection}

\begin{figure*}[t]\centering
\includegraphics[width=0.91\linewidth]{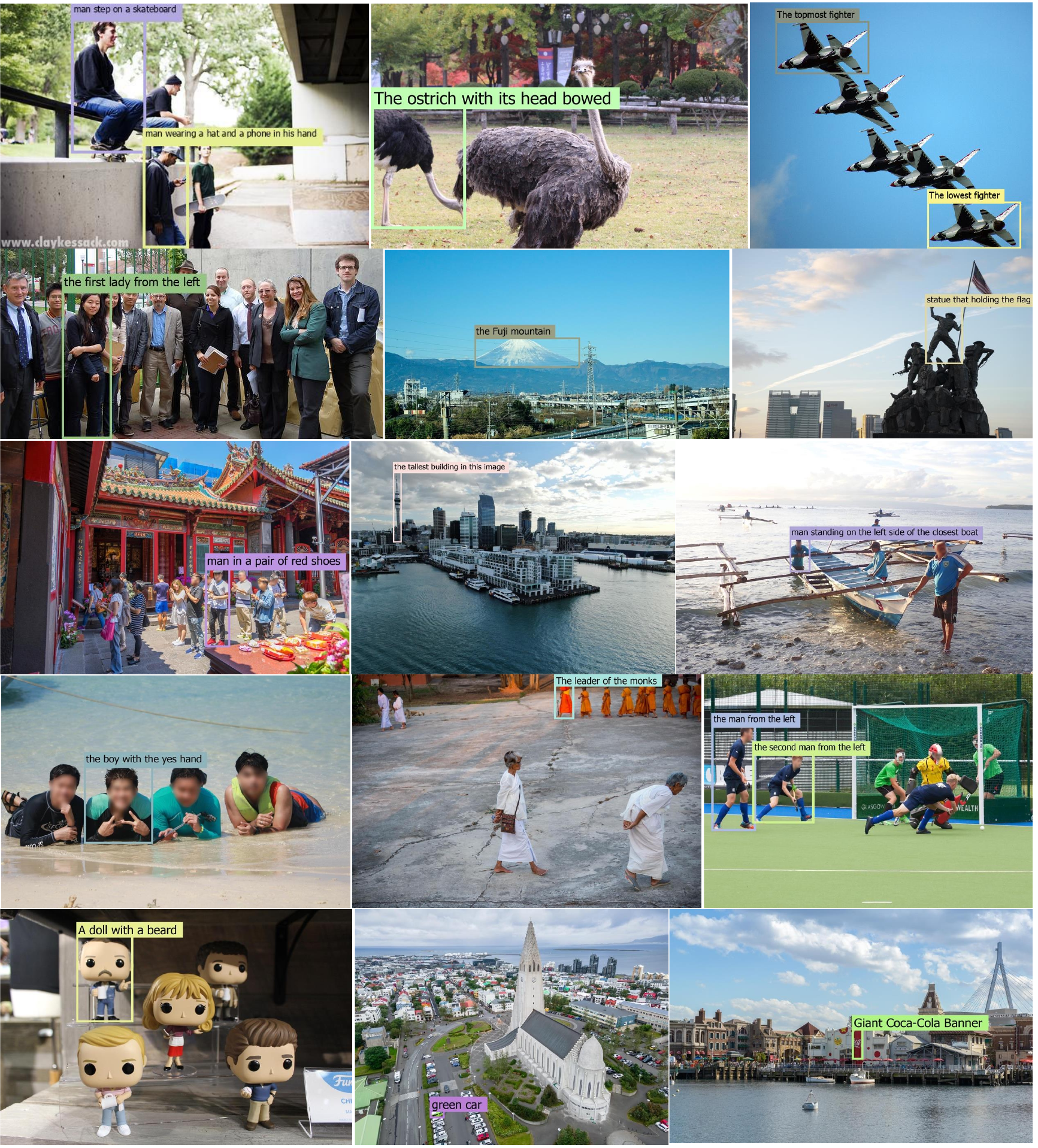}\vspace{-1mm}
\caption{Visualization on the Referring Object Detection Task.}
\label{fig:vis_referring_object_det}
\vspace{-1mm}
\end{figure*}

We show the results on the referring object detection task in Fig. \ref{fig:vis_referring_object_det}.

\subsection{Region Caption}

\begin{figure*}[t]\centering
\includegraphics[width=0.91\linewidth]{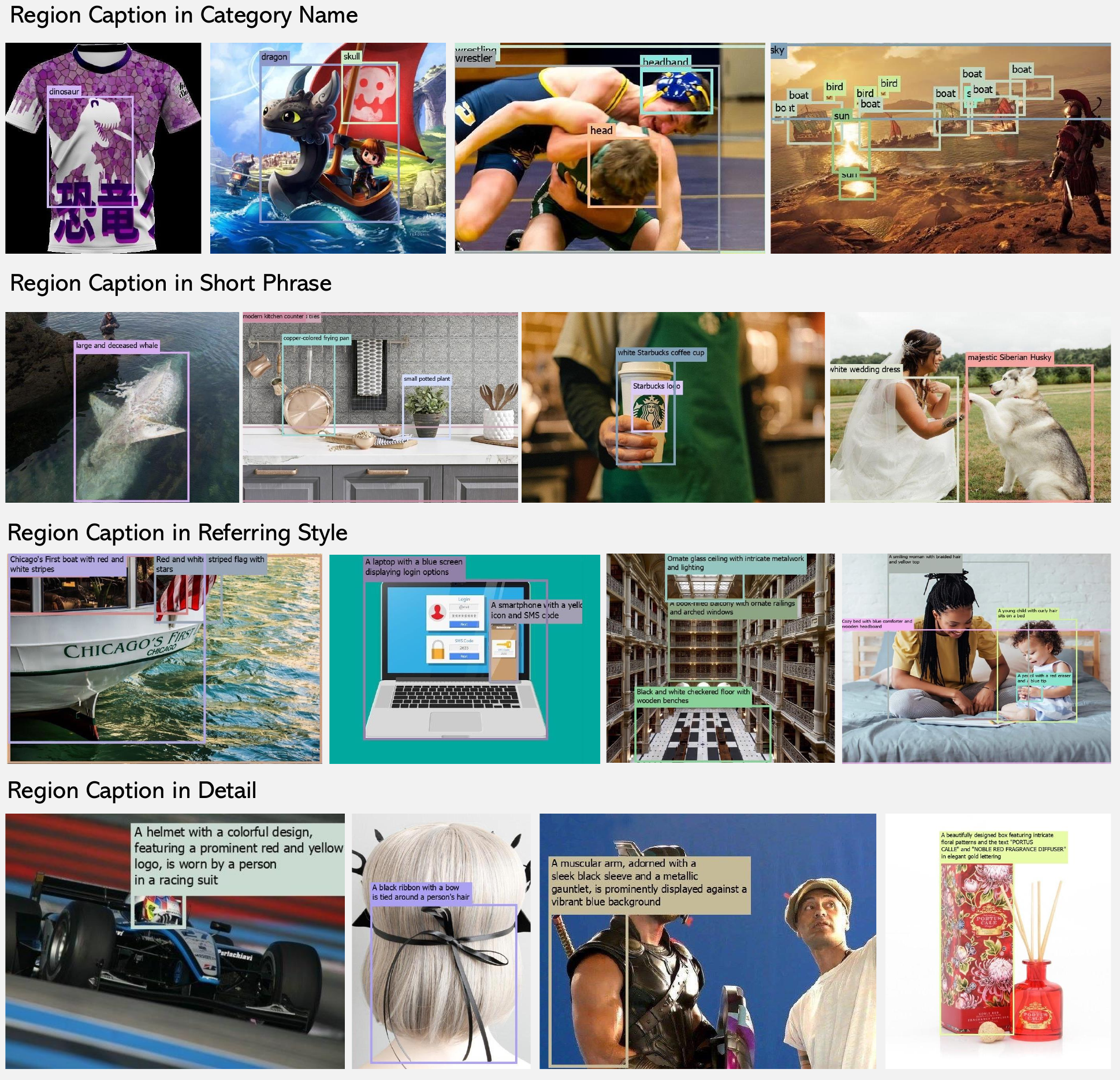}\vspace{-1mm}
\caption{Visualization on Region Caption Tasks.}
\label{fig:vis_region_caption}
\vspace{-1mm}
\end{figure*}

We show the results on the region caption task in Fig. \ref{fig:vis_region_caption}.

\subsection{Region QA}

\begin{figure*}[t]\centering
\includegraphics[width=0.91\linewidth]{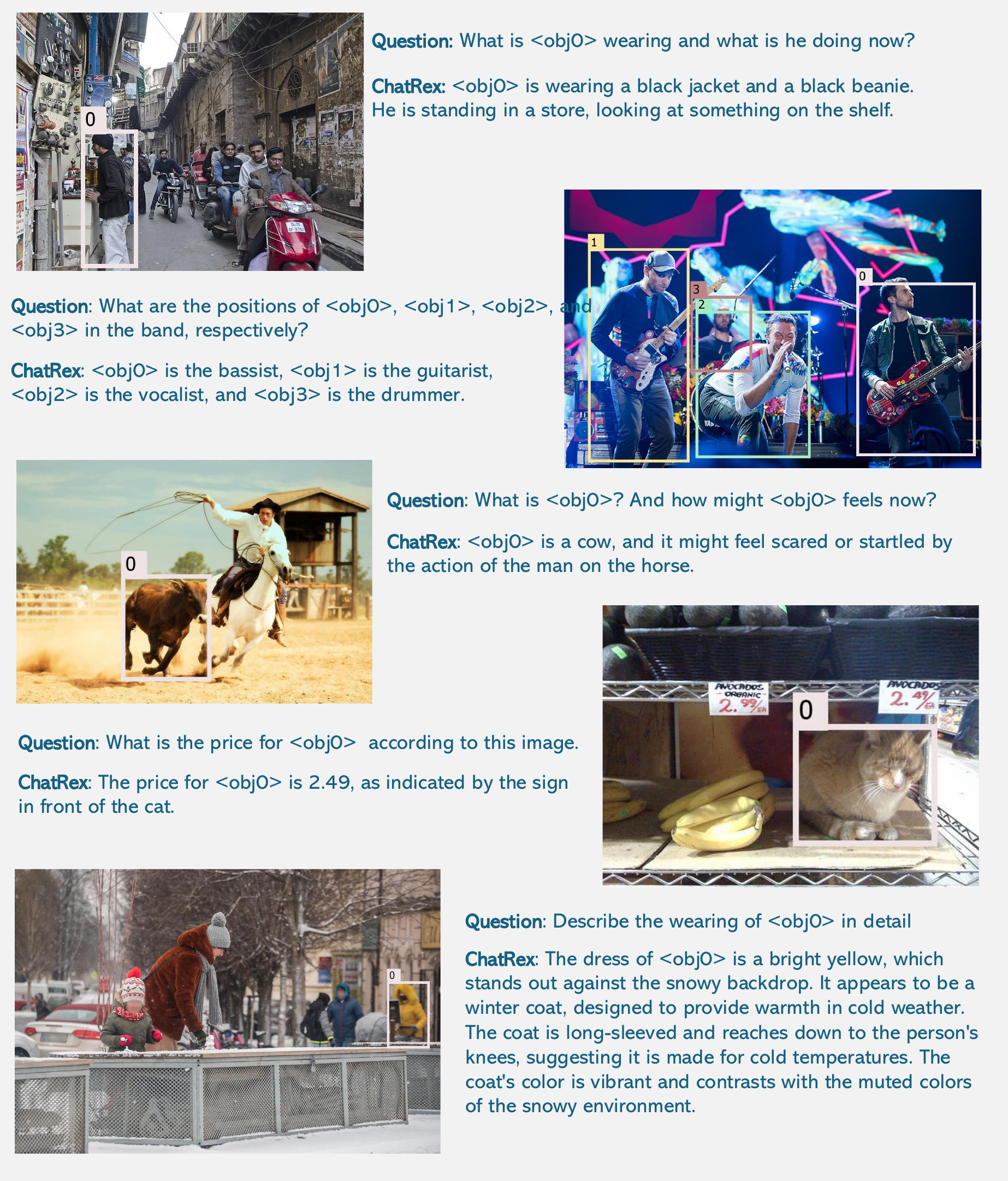}\vspace{-1mm}
\caption{Visualization on the Region QA Task.}
\label{fig:vis_region_qa}
\vspace{-1mm}
\end{figure*}

We show the results on the region QA task in Fig. \ref{fig:vis_region_qa}.

\subsection{Brief Grounded Image Caption}

\begin{figure*}[t]\centering
\includegraphics[width=0.91\linewidth]{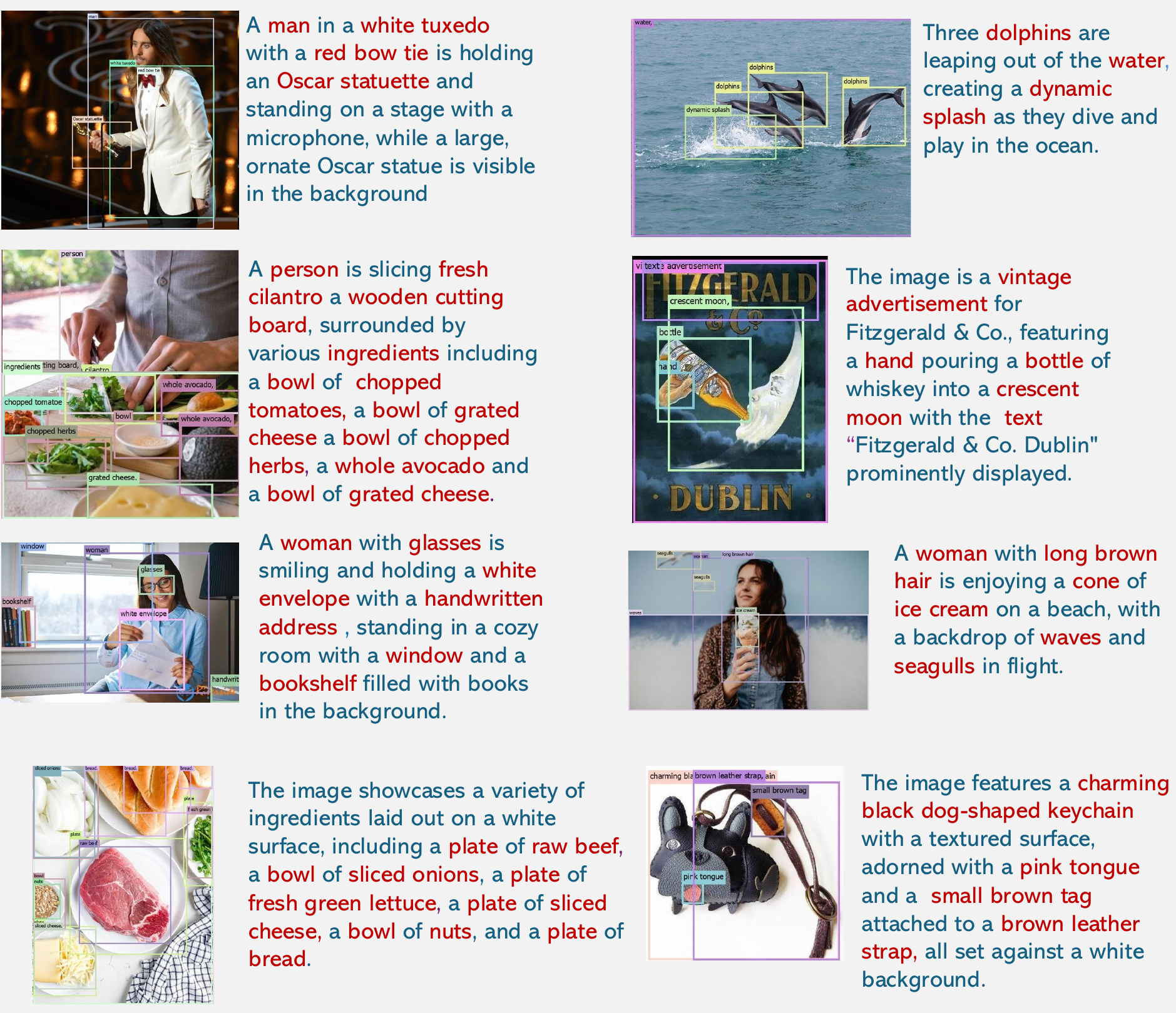}\vspace{-1mm}
\caption{Visualization on the Brief Grounded Image Caption Task.}
\label{fig:vis_breif_grounded_caption}
\vspace{-1mm}
\end{figure*}

We show the results on the brief grounded image caption task in Fig. \ref{fig:vis_breif_grounded_caption}.

\subsection{Detailed Grounded Image Caption}

\begin{figure*}[t]\centering
\includegraphics[width=0.91\linewidth]{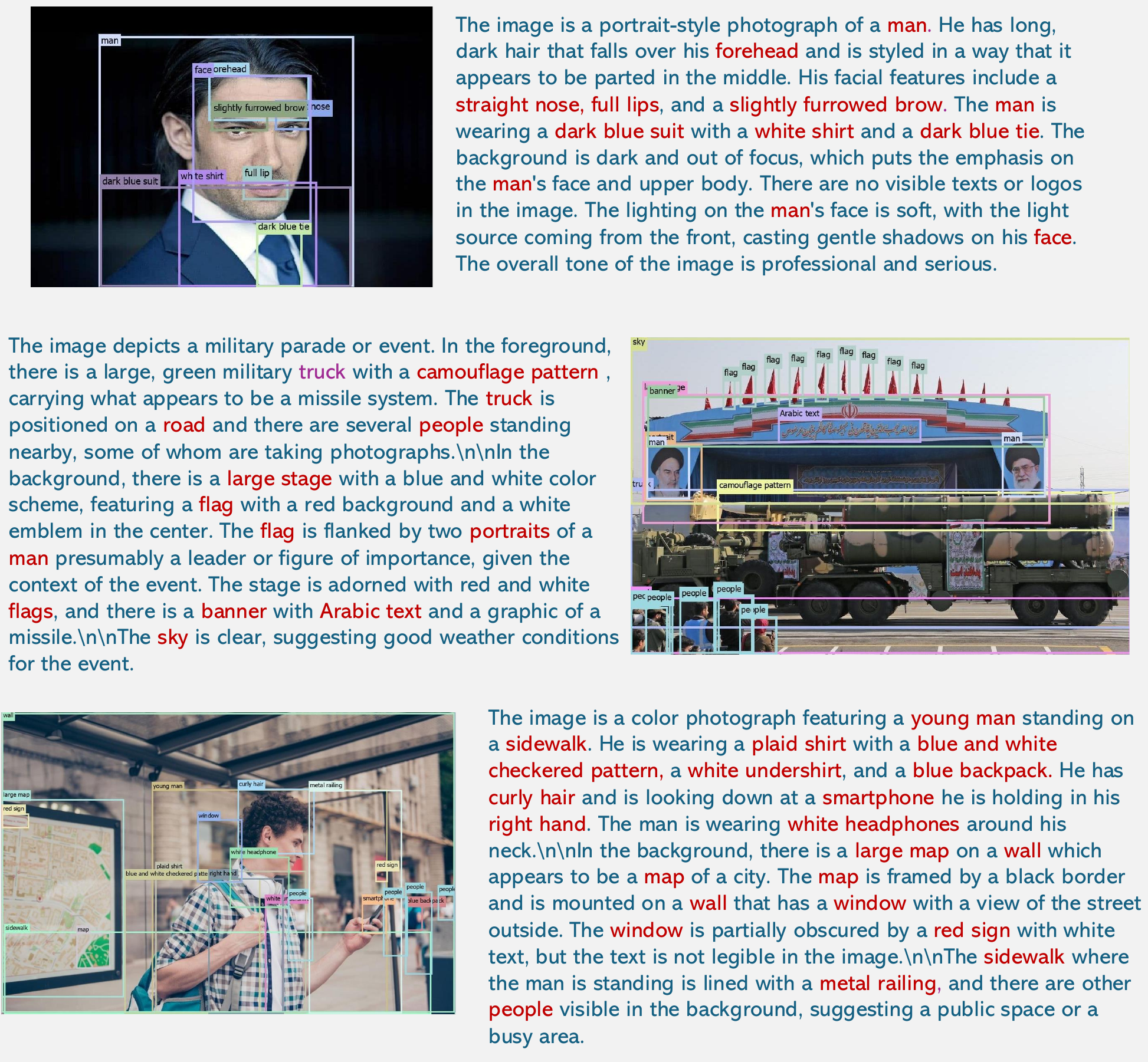}\vspace{-1mm}
\caption{Visualization on the Detailed Grounded Image Caption Task.}
\label{fig:vis_detailed_grounded_caption}
\vspace{-1mm}
\end{figure*}

We show the results on the detailed grounded image caption task in Fig. \ref{fig:vis_detailed_grounded_caption}.

\subsection{Grounded Counting}

\begin{figure*}[t]\centering
\includegraphics[width=0.91\linewidth]{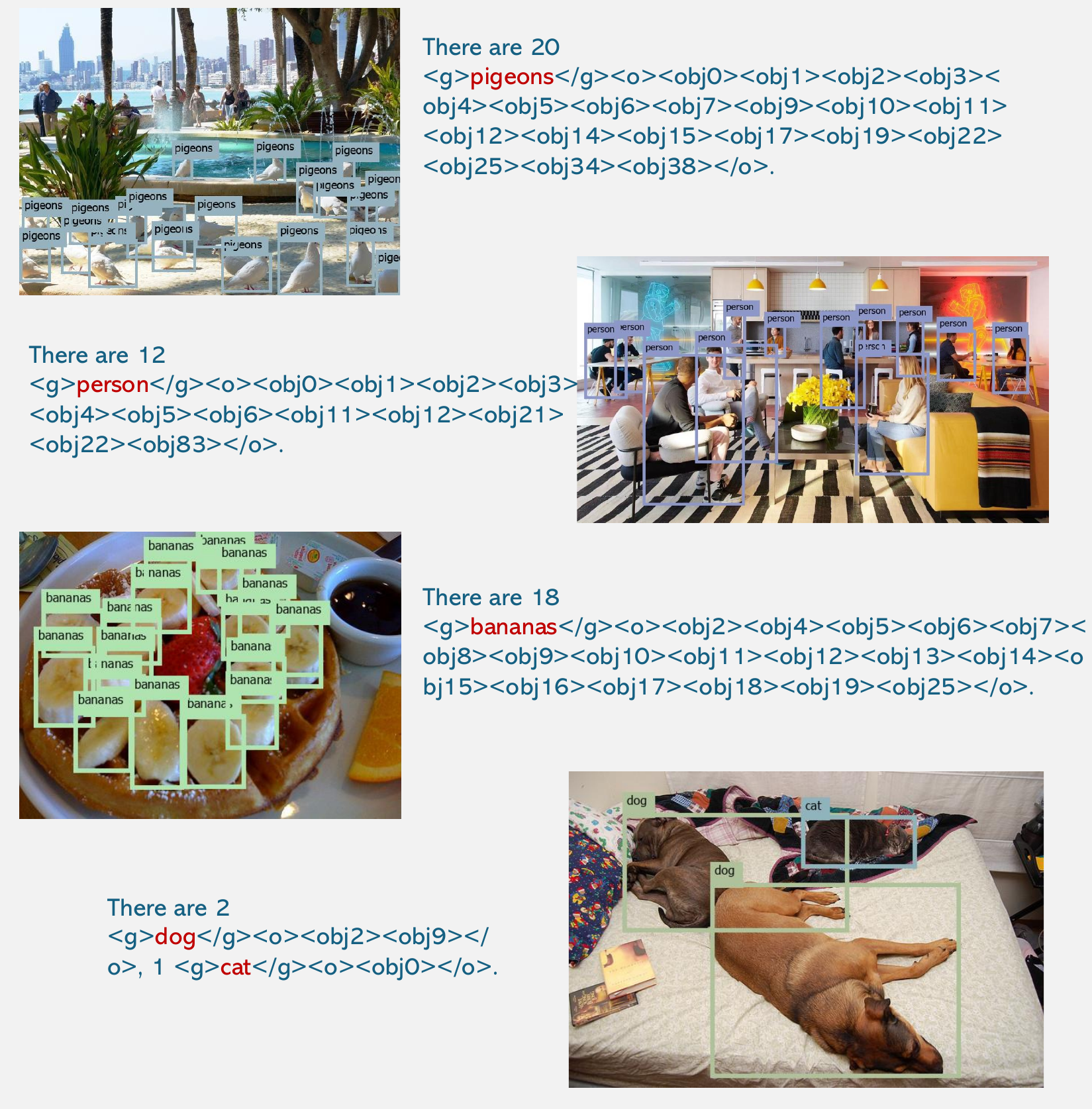}\vspace{-1mm}
\caption{Visualization on the Grounded Counting Task.}
\label{fig:vis_grounded_counting}
\vspace{-1mm}
\end{figure*}

We show the results on the grounded counting task in Fig. \ref{fig:vis_grounded_counting}.

\subsection{Grounded Conversation}

\begin{figure*}[t]\centering
\includegraphics[width=0.91\linewidth]{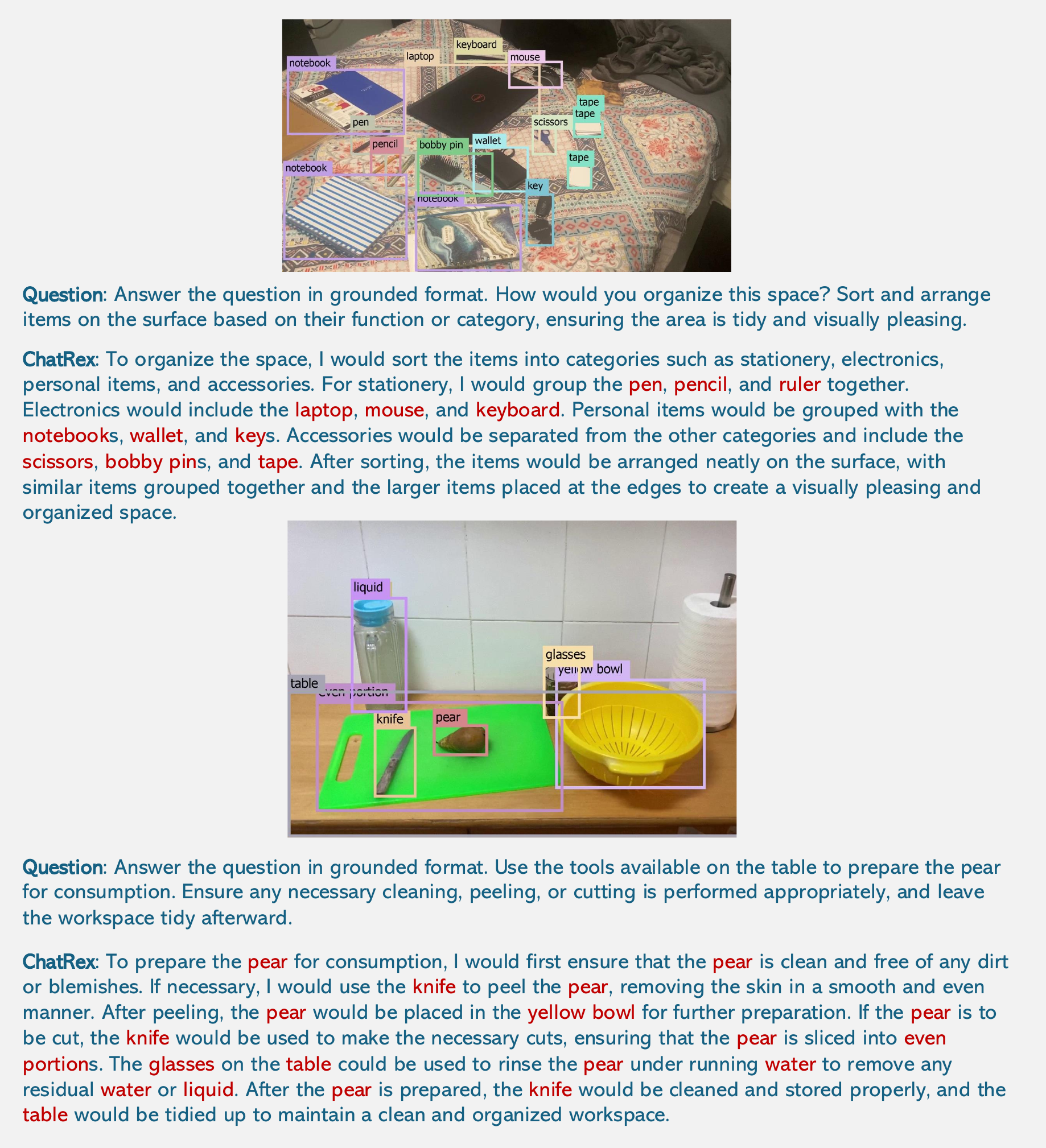}\vspace{-1mm}
\caption{Visualization on the Grounded Counting Task.}
\label{fig:vis_grounded_conv1}
\vspace{-1mm}
\end{figure*}

\begin{figure*}[t]\centering
\includegraphics[width=0.91\linewidth]{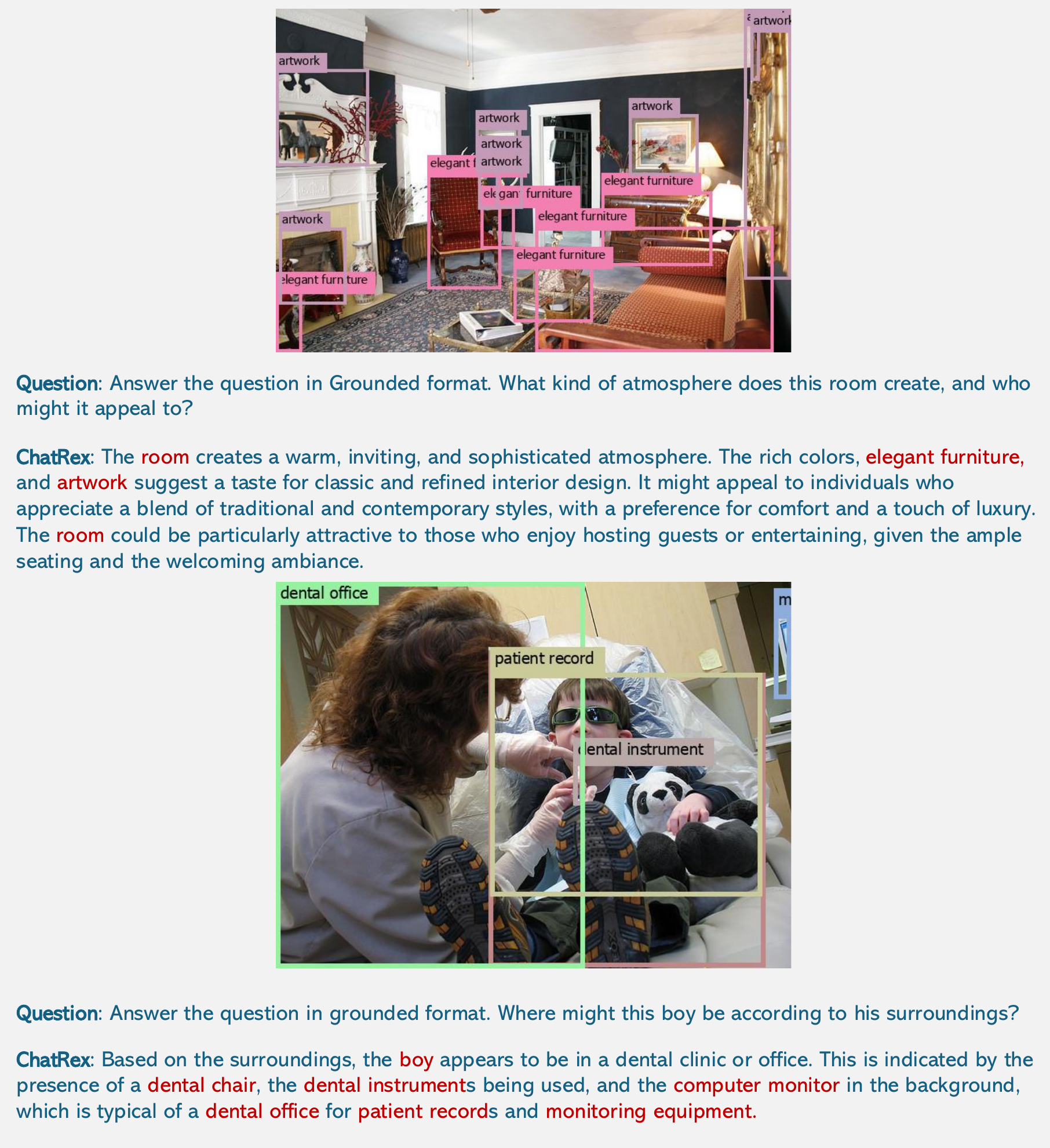}\vspace{-1mm}
\caption{Visualization on the Grounded Counting Task.}
\label{fig:vis_grounded_conv2}
\vspace{-1mm}
\end{figure*}

We show the results on the grounded conversation task in Fig. \ref{fig:vis_grounded_conv1} and Fig. \ref{fig:vis_grounded_conv2}.

%% file: main.bbl
\begin{thebibliography}{109}
\providecommand{\natexlab}[1]{#1}
\providecommand{\url}[1]{\texttt{#1}}
\expandafter\ifx\csname urlstyle\endcsname\relax
  \providecommand{\doi}[1]{doi: #1}\else
  \providecommand{\doi}{doi: \begingroup \urlstyle{rm}\Url}\fi

\bibitem[Abdin et~al.(2024)Abdin, Aneja, Awadalla, Awadallah, Awan, Bach, Bahree, Bakhtiari, Bao, Behl, et~al.]{abdin2024phi}
Marah Abdin, Jyoti Aneja, Hany Awadalla, Ahmed Awadallah, Ammar~Ahmad Awan, Nguyen Bach, Amit Bahree, Arash Bakhtiari, Jianmin Bao, Harkirat Behl, et~al.
\newblock Phi-3 technical report: A highly capable language model locally on your phone.
\newblock \emph{arXiv preprint arXiv:2404.14219}, 2024.

\bibitem[Achiam et~al.(2023)Achiam, Adler, Agarwal, Ahmad, Akkaya, Aleman, Almeida, Altenschmidt, Altman, Anadkat, et~al.]{openai2023gpt4}
Josh Achiam, Steven Adler, Sandhini Agarwal, Lama Ahmad, Ilge Akkaya, Florencia~Leoni Aleman, Diogo Almeida, Janko Altenschmidt, Sam Altman, Shyamal Anadkat, et~al.
\newblock Gpt-4 technical report.
\newblock \emph{arXiv preprint arXiv:2303.08774}, 2023.

\bibitem[Agrawal et~al.(2024)Agrawal, Antoniak, Hanna, Chaplot, Chudnovsky, Garg, Gervet, Ghosh, H{\'e}liou, Jacob, et~al.]{agrawal2024pixtral}
Pravesh Agrawal, Szymon Antoniak, Emma~Bou Hanna, Devendra Chaplot, Jessica Chudnovsky, Saurabh Garg, Theophile Gervet, Soham Ghosh, Am{\'e}lie H{\'e}liou, Paul Jacob, et~al.
\newblock Pixtral 12b.
\newblock \emph{arXiv preprint arXiv:2410.07073}, 2024.

\bibitem[Alayrac et~al.(2022)Alayrac, Donahue, Luc, Miech, Barr, Hasson, Lenc, Mensch, Millican, Reynolds, et~al.]{alayrac2022flamingo}
Jean-Baptiste Alayrac, Jeff Donahue, Pauline Luc, Antoine Miech, Iain Barr, Yana Hasson, Karel Lenc, Arthur Mensch, Katherine Millican, Malcolm Reynolds, et~al.
\newblock Flamingo: a visual language model for few-shot learning.
\newblock \emph{NeurIPS}, 35:\penalty0 23716--23736, 2022.

\bibitem[Bai et~al.(2023{\natexlab{a}})Bai, Bai, Chu, Cui, Dang, Deng, Fan, Ge, Han, Huang, Hui, Ji, Li, Lin, Lin, Liu, Liu, Lu, Lu, Ma, Men, Ren, Ren, Tan, Tan, Tu, Wang, Wang, Wang, Wu, Xu, Xu, Yang, Yang, Yang, Yang, Yao, Yu, Yuan, Yuan, Zhang, Zhang, Zhang, Zhang, Zhou, Zhou, Zhou, and Zhu]{qwen}
Jinze Bai, Shuai Bai, Yunfei Chu, Zeyu Cui, Kai Dang, Xiaodong Deng, Yang Fan, Wenbin Ge, Yu Han, Fei Huang, Binyuan Hui, Luo Ji, Mei Li, Junyang Lin, Runji Lin, Dayiheng Liu, Gao Liu, Chengqiang Lu, Keming Lu, Jianxin Ma, Rui Men, Xingzhang Ren, Xuancheng Ren, Chuanqi Tan, Sinan Tan, Jianhong Tu, Peng Wang, Shijie Wang, Wei Wang, Shengguang Wu, Benfeng Xu, Jin Xu, An Yang, Hao Yang, Jian Yang, Shusheng Yang, Yang Yao, Bowen Yu, Hongyi Yuan, Zheng Yuan, Jianwei Zhang, Xingxuan Zhang, Yichang Zhang, Zhenru Zhang, Chang Zhou, Jingren Zhou, Xiaohuan Zhou, and Tianhang Zhu.
\newblock Qwen technical report.
\newblock \emph{arXiv preprint arXiv:2309.16609}, 2023{\natexlab{a}}.

\bibitem[Bai et~al.(2023{\natexlab{b}})Bai, Bai, Yang, Wang, Tan, Wang, Lin, Zhou, and Zhou]{bai2023qwenvl}
Jinze Bai, Shuai Bai, Shusheng Yang, Shijie Wang, Sinan Tan, Peng Wang, Junyang Lin, Chang Zhou, and Jingren Zhou.
\newblock Qwen-vl: A frontier large vision-language model with versatile abilities.
\newblock \emph{arXiv preprint arXiv:2308.12966}, 2023{\natexlab{b}}.

\bibitem[Byeon et~al.(2022)Byeon, Park, Kim, Lee, Baek, and Kim]{kakaobrain2022coyo-700m}
Minwoo Byeon, Beomhee Park, Haecheon Kim, Sungjun Lee, Woonhyuk Baek, and Saehoon Kim.
\newblock Coyo-700m: Image-text pair dataset.
\newblock \url{https://github.com/kakaobrain/coyo-dataset}, 2022.

\bibitem[Carion et~al.(2020)Carion, Massa, Synnaeve, Usunier, Kirillov, and Zagoruyko]{carion2020end}
Nicolas Carion, Francisco Massa, Gabriel Synnaeve, Nicolas Usunier, Alexander Kirillov, and Sergey Zagoruyko.
\newblock End-to-end object detection with transformers.
\newblock In \emph{European conference on computer vision}, pages 213--229. Springer, 2020.

\bibitem[Chen et~al.(2024{\natexlab{a}})Chen, Chen, Zhang, Chen, Wu, Zhang, Chen, Li, Wan, and Wang]{chen2024allava}
Guiming~Hardy Chen, Shunian Chen, Ruifei Zhang, Junying Chen, Xiangbo Wu, Zhiyi Zhang, Zhihong Chen, Jianquan Li, Xiang Wan, and Benyou Wang.
\newblock Allava: Harnessing gpt4v-synthesized data for a lite vision-language model.
\newblock \emph{arXiv preprint arXiv:2402.11684}, 2024{\natexlab{a}}.

\bibitem[Chen et~al.(2023)Chen, Zhang, Zeng, Zhang, Zhu, and Zhao]{chen2023shikra}
Keqin Chen, Zhao Zhang, Weili Zeng, Richong Zhang, Feng Zhu, and Rui Zhao.
\newblock Shikra: Unleashing multimodal llm's referential dialogue magic.
\newblock \emph{arXiv preprint arXiv:2306.15195}, 2023.

\bibitem[Chen et~al.(2024{\natexlab{b}})Chen, Li, Dong, Zhang, Zang, Chen, Duan, Wang, Qiao, Lin, et~al.]{chen2024we}
Lin Chen, Jinsong Li, Xiaoyi Dong, Pan Zhang, Yuhang Zang, Zehui Chen, Haodong Duan, Jiaqi Wang, Yu Qiao, Dahua Lin, et~al.
\newblock Are we on the right way for evaluating large vision-language models?
\newblock \emph{arXiv preprint arXiv:2403.20330}, 2024{\natexlab{b}}.

\bibitem[Chen et~al.(2021)Chen, Saxena, Li, Fleet, and Hinton]{chen2021pix2seq}
Ting Chen, Saurabh Saxena, Lala Li, David~J Fleet, and Geoffrey Hinton.
\newblock Pix2seq: A language modeling framework for object detection.
\newblock \emph{arXiv preprint arXiv:2109.10852}, 2021.

\bibitem[Chen et~al.(2022)Chen, Wang, Changpinyo, Piergiovanni, Padlewski, Salz, Goodman, Grycner, Mustafa, Beyer, et~al.]{chen2022pali}
Xi Chen, Xiao Wang, Soravit Changpinyo, AJ Piergiovanni, Piotr Padlewski, Daniel Salz, Sebastian Goodman, Adam Grycner, Basil Mustafa, Lucas Beyer, et~al.
\newblock Pali: A jointly-scaled multilingual language-image model.
\newblock In \emph{ICLR}, 2022.

\bibitem[Chen et~al.(2024{\natexlab{c}})Chen, Wang, Tian, Ye, Gao, Cui, Tong, Hu, Luo, Ma, et~al.]{VLM:InternVL-1.5}
Zhe Chen, Weiyun Wang, Hao Tian, Shenglong Ye, Zhangwei Gao, Erfei Cui, Wenwen Tong, Kongzhi Hu, Jiapeng Luo, Zheng Ma, et~al.
\newblock How far are we to gpt-4v? closing the gap to commercial multimodal models with open-source suites.
\newblock \emph{arXiv:2404.16821}, 2024{\natexlab{c}}.

\bibitem[Chiang et~al.(2023)Chiang, Li, Lin, Sheng, Wu, Zhang, Zheng, Zhuang, Zhuang, Gonzalez, Stoica, and Xing]{TransF:Vicuna}
Wei-Lin Chiang, Zhuohan Li, Zi Lin, Ying Sheng, Zhanghao Wu, Hao Zhang, Lianmin Zheng, Siyuan Zhuang, Yonghao Zhuang, Joseph~E. Gonzalez, Ion Stoica, and Eric~P. Xing.
\newblock Vicuna: An open-source chatbot impressing gpt-4 with 90\%* chatgpt quality, 2023.

\bibitem[Dai et~al.(2023)Dai, Li, Li, Tiong, Zhao, Wang, Li, Fung, and Hoi]{VLM:InstructBLIP}
Wenliang Dai, Junnan Li, Dongxu Li, Anthony Meng~Huat Tiong, Junqi Zhao, Weisheng Wang, Boyang Li, Pascale Fung, and Steven C.~H. Hoi.
\newblock Instructblip: Towards general-purpose vision-language models with instruction tuning.
\newblock In \emph{NeurIPS}, 2023.

\bibitem[Dai et~al.(2024)Dai, Lee, Wang, Yang, Liu, Barker, Rintamaki, Shoeybi, Catanzaro, and Ping]{dai2024nvlm}
Wenliang Dai, Nayeon Lee, Boxin Wang, Zhuoling Yang, Zihan Liu, Jon Barker, Tuomas Rintamaki, Mohammad Shoeybi, Bryan Catanzaro, and Wei Ping.
\newblock Nvlm: Open frontier-class multimodal llms.
\newblock \emph{arXiv preprint arXiv:2409.11402}, 2024.

\bibitem[Deitke et~al.(2024)Deitke, Clark, Lee, Tripathi, Yang, Park, Salehi, Muennighoff, Lo, Soldaini, et~al.]{deitke2024molmo}
Matt Deitke, Christopher Clark, Sangho Lee, Rohun Tripathi, Yue Yang, Jae~Sung Park, Mohammadreza Salehi, Niklas Muennighoff, Kyle Lo, Luca Soldaini, et~al.
\newblock Molmo and pixmo: Open weights and open data for state-of-the-art multimodal models.
\newblock \emph{arXiv preprint arXiv:2409.17146}, 2024.

\bibitem[Deng et~al.(2009)Deng, Dong, Socher, Li, Li, and Fei-Fei]{deng2009imagenet}
Jia Deng, Wei Dong, Richard Socher, Li-Jia Li, Kai Li, and Li Fei-Fei.
\newblock Imagenet: A large-scale hierarchical image database.
\newblock In \emph{CVPR}, pages 248--255, 2009.

\bibitem[Dong et~al.(2024)Dong, Zhang, Zang, Cao, Wang, Ouyang, Zhang, Duan, Zhang, Li, Yan, Gao, Chen, Zhang, Li, Li, Wang, Chen, He, Zhang, Dai, Qiao, Lin, and Wang]{VLM:Xcomposer2-4KHD}
Xiaoyi Dong, Pan Zhang, Yuhang Zang, Yuhang Cao, Bin Wang, Linke Ouyang, Songyang Zhang, Haodong Duan, Wenwei Zhang, Yining Li, Hang Yan, Yang Gao, Zhe Chen, Xinyue Zhang, Wei Li, Jingwen Li, Wenhai Wang, Kai Chen, Conghui He, Xingcheng Zhang, Jifeng Dai, Yu Qiao, Dahua Lin, and Jiaqi Wang.
\newblock Internlm-xcomposer2-4khd: {A} pioneering large vision-language model handling resolutions from 336 pixels to 4k {HD}.
\newblock \emph{arXiv: 2404.06512}, 2024.

\bibitem[Dosovitskiy et~al.(2021)Dosovitskiy, Beyer, Kolesnikov, Weissenborn, Zhai, Unterthiner, Dehghani, Minderer, Heigold, Gelly, Uszkoreit, and Houlsby]{TransF:ViT}
Alexey Dosovitskiy, Lucas Beyer, Alexander Kolesnikov, Dirk Weissenborn, Xiaohua Zhai, Thomas Unterthiner, Mostafa Dehghani, Matthias Minderer, Georg Heigold, Sylvain Gelly, Jakob Uszkoreit, and Neil Houlsby.
\newblock An image is worth 16x16 words: Transformers for image recognition at scale.
\newblock In \emph{ICLR}, 2021.

\bibitem[Dubey et~al.(2024)Dubey, Jauhri, Pandey, Kadian, Al-Dahle, Letman, Mathur, Schelten, Yang, Fan, et~al.]{dubey2024llama}
Abhimanyu Dubey, Abhinav Jauhri, Abhinav Pandey, Abhishek Kadian, Ahmad Al-Dahle, Aiesha Letman, Akhil Mathur, Alan Schelten, Amy Yang, Angela Fan, et~al.
\newblock The llama 3 herd of models.
\newblock \emph{arXiv preprint arXiv:2407.21783}, 2024.

\bibitem[Fu et~al.(2024)Fu, Chen, Shen, Qin, Zhang, Lin, Yang, Zheng, Li, Sun, Wu, and Ji]{fu2024mmecomprehensiveevaluationbenchmark}
Chaoyou Fu, Peixian Chen, Yunhang Shen, Yulei Qin, Mengdan Zhang, Xu Lin, Jinrui Yang, Xiawu Zheng, Ke Li, Xing Sun, Yunsheng Wu, and Rongrong Ji.
\newblock Mme: A comprehensive evaluation benchmark for multimodal large language models, 2024.

\bibitem[Guan et~al.(2023)Guan, Liu, Wu, Xian, Li, Liu, Wang, Chen, Huang, Yacoob, et~al.]{guan2023hallusionbench}
Tianrui Guan, Fuxiao Liu, Xiyang Wu, Ruiqi Xian, Zongxia Li, Xiaoyu Liu, Xijun Wang, Lichang Chen, Furong Huang, Yaser Yacoob, et~al.
\newblock Hallusionbench: An advanced diagnostic suite for entangled language hallucination \& visual illusion in large vision-language models.
\newblock \emph{arXiv preprint arXiv:2310.14566}, 2023.

\bibitem[Gupta et~al.(2019)Gupta, Dollar, and Girshick]{gupta2019lvis}
Agrim Gupta, Piotr Dollar, and Ross Girshick.
\newblock Lvis: A dataset for large vocabulary instance segmentation.
\newblock In \emph{Proceedings of the IEEE/CVF conference on computer vision and pattern recognition}, pages 5356--5364, 2019.

\bibitem[He et~al.(2017)He, Gkioxari, Doll{\'a}r, and Girshick]{he2017mask}
Kaiming He, Georgia Gkioxari, Piotr Doll{\'a}r, and Ross Girshick.
\newblock Mask r-cnn.
\newblock In \emph{ICCV}, pages 2961--2969, 2017.

\bibitem[Huang et~al.(2019)Huang, Chen, He, Bai, Karatzas, Lu, and Jawahar]{sroie}
Zheng Huang, Kai Chen, Jianhua He, Xiang Bai, Dimosthenis Karatzas, Shijian Lu, and CV Jawahar.
\newblock Icdar2019 competition on scanned receipt ocr and information extraction.
\newblock In \emph{2019 International Conference on Document Analysis and Recognition (ICDAR)}, pages 1516--1520. IEEE, 2019.

\bibitem[Jiang et~al.(2024)Jiang, He, Zeng, Wei, Ku, Liu, and Chen]{VLM:MANTIS}
Dongfu Jiang, Xuan He, Huaye Zeng, Con Wei, Max Ku, Qian Liu, and Wenhu Chen.
\newblock Mantis: Interleaved multi-image instruction tuning.
\newblock \emph{arXiv:2405.01483}, 2024.

\bibitem[Jiang et~al.(2025)Jiang, Li, Zeng, Ren, Liu, and Zhang]{jiang2025t}
Qing Jiang, Feng Li, Zhaoyang Zeng, Tianhe Ren, Shilong Liu, and Lei Zhang.
\newblock T-rex2: Towards generic object detection via text-visual prompt synergy.
\newblock In \emph{European Conference on Computer Vision}, pages 38--57. Springer, 2025.

\bibitem[Kazemzadeh et~al.(2014)Kazemzadeh, Ordonez, Matten, and Berg]{kazemzadeh2014referitgame}
Sahar Kazemzadeh, Vicente Ordonez, Mark Matten, and Tamara Berg.
\newblock Referitgame: Referring to objects in photographs of natural scenes.
\newblock In \emph{Proceedings of the 2014 conference on empirical methods in natural language processing (EMNLP)}, pages 787--798, 2014.

\bibitem[Kembhavi et~al.(2016)Kembhavi, Salvato, Kolve, Seo, Hajishirzi, and Farhadi]{kembhavi2016diagram}
Aniruddha Kembhavi, Mike Salvato, Eric Kolve, Minjoon Seo, Hannaneh Hajishirzi, and Ali Farhadi.
\newblock A diagram is worth a dozen images.
\newblock In \emph{Computer Vision--ECCV 2016: 14th European Conference, Amsterdam, The Netherlands, October 11--14, 2016, Proceedings, Part IV 14}, pages 235--251. Springer, 2016.

\bibitem[Kirillov et~al.(2023)Kirillov, Mintun, Ravi, Mao, Rolland, Gustafson, Xiao, Whitehead, Berg, Lo, Doll{\'{a}}r, and Girshick]{TransF:SAM}
Alexander Kirillov, Eric Mintun, Nikhila Ravi, Hanzi Mao, Chlo{\'{e}} Rolland, Laura Gustafson, Tete Xiao, Spencer Whitehead, Alexander~C. Berg, Wan{-}Yen Lo, Piotr Doll{\'{a}}r, and Ross~B. Girshick.
\newblock Segment anything.
\newblock \emph{arXiv: 2304.02643}, 2023.

\bibitem[Kuznetsova et~al.(2020)Kuznetsova, Rom, Alldrin, Uijlings, Krasin, Pont-Tuset, Kamali, Popov, Malloci, Kolesnikov, et~al.]{openimages}
Alina Kuznetsova, Hassan Rom, Neil Alldrin, Jasper Uijlings, Ivan Krasin, Jordi Pont-Tuset, Shahab Kamali, Stefan Popov, Matteo Malloci, Alexander Kolesnikov, et~al.
\newblock The open images dataset v4: Unified image classification, object detection, and visual relationship detection at scale.
\newblock \emph{International journal of computer vision}, 128\penalty0 (7):\penalty0 1956--1981, 2020.

\bibitem[Lai et~al.(2023)Lai, Tian, Chen, Li, Yuan, Liu, and Jia]{lai2023lisa}
Xin Lai, Zhuotao Tian, Yukang Chen, Yanwei Li, Yuhui Yuan, Shu Liu, and Jiaya Jia.
\newblock Lisa: Reasoning segmentation via large language model.
\newblock \emph{arXiv preprint arXiv:2308.00692}, 2023.

\bibitem[Li et~al.(2023{\natexlab{a}})Li, Wang, Wang, Ge, Ge, and Shan]{li2023seed}
Bohao Li, Rui Wang, Guangzhi Wang, Yuying Ge, Yixiao Ge, and Ying Shan.
\newblock Seed-bench: Benchmarking multimodal llms with generative comprehension.
\newblock \emph{arXiv preprint arXiv:2307.16125}, 2023{\natexlab{a}}.

\bibitem[Li et~al.(2024{\natexlab{a}})Li, Liu, Wu, Wang, Shen, Qu, Niu, Wang, Chen, and Li]{li2024aria}
Dongxu Li, Yudong Liu, Haoning Wu, Yue Wang, Zhiqi Shen, Bowen Qu, Xinyao Niu, Guoyin Wang, Bei Chen, and Junnan Li.
\newblock Aria: An open multimodal native mixture-of-experts model.
\newblock \emph{arXiv preprint arXiv:2410.05993}, 2024{\natexlab{a}}.

\bibitem[Li et~al.(2024{\natexlab{b}})Li, Zhang, Zhang, Zhang, Li, Li, Ma, and Li]{li2024llava}
Feng Li, Renrui Zhang, Hao Zhang, Yuanhan Zhang, Bo Li, Wei Li, Zejun Ma, and Chunyuan Li.
\newblock Llava-next-interleave: Tackling multi-image, video, and 3d in large multimodal models.
\newblock \emph{arXiv preprint arXiv:2407.07895}, 2024{\natexlab{b}}.

\bibitem[Li et~al.(2023{\natexlab{b}})Li, Li, Savarese, and Hoi]{li2023blip2}
Junnan Li, Dongxu Li, Silvio Savarese, and Steven Hoi.
\newblock Blip-2: Bootstrapping language-image pre-training with frozen image encoders and large language models.
\newblock In \emph{ICML}, pages 19730--19742. PMLR, 2023{\natexlab{b}}.

\bibitem[Li et~al.(2022)Li, Zhang, Zhang, Yang, Li, Zhong, Wang, Yuan, Zhang, Hwang, et~al.]{li2022grounded}
Liunian~Harold Li, Pengchuan Zhang, Haotian Zhang, Jianwei Yang, Chunyuan Li, Yiwu Zhong, Lijuan Wang, Lu Yuan, Lei Zhang, Jenq-Neng Hwang, et~al.
\newblock Grounded language-image pre-training.
\newblock In \emph{Proceedings of the IEEE/CVF Conference on Computer Vision and Pattern Recognition}, pages 10965--10975, 2022.

\bibitem[Li et~al.(2023{\natexlab{c}})Li, Du, Zhou, Wang, Zhao, and Wen]{Datasets:POPE}
Yifan Li, Yifan Du, Kun Zhou, Jinpeng Wang, Wayne~Xin Zhao, and Ji{-}Rong Wen.
\newblock Evaluating object hallucination in large vision-language models.
\newblock In \emph{EMNLP}, pages 292--305, 2023{\natexlab{c}}.

\bibitem[Li et~al.(2024{\natexlab{c}})Li, Zhang, Wang, Zhong, Chen, Chu, Liu, and Jia]{VLM:MiniGemini}
Yanwei Li, Yuechen Zhang, Chengyao Wang, Zhisheng Zhong, Yixin Chen, Ruihang Chu, Shaoteng Liu, and Jiaya Jia.
\newblock Mini-gemini: Mining the potential of multi-modality vision language models.
\newblock \emph{arXiv: 2403.18814}, 2024{\natexlab{c}}.

\bibitem[Li et~al.(2024{\natexlab{d}})Li, Zhang, Wang, Zhong, Chen, Chu, Liu, and Jia]{li2024mini}
Yanwei Li, Yuechen Zhang, Chengyao Wang, Zhisheng Zhong, Yixin Chen, Ruihang Chu, Shaoteng Liu, and Jiaya Jia.
\newblock Mini-gemini: Mining the potential of multi-modality vision language models.
\newblock \emph{arXiv preprint arXiv:2403.18814}, 2024{\natexlab{d}}.

\bibitem[Li et~al.(2023{\natexlab{d}})Li, Yang, Liu, Ma, Zhang, Yang, Sun, Liu, and Bai]{li2023monkey}
Zhang Li, Biao Yang, Qiang Liu, Zhiyin Ma, Shuo Zhang, Jingxu Yang, Yabo Sun, Yuliang Liu, and Xiang Bai.
\newblock Monkey: Image resolution and text label are important things for large multi-modal models.
\newblock \emph{arXiv preprint arXiv:2311.06607}, 2023{\natexlab{d}}.

\bibitem[Lin et~al.(2023{\natexlab{a}})Lin, Ye, Zhu, Cui, Ning, Jin, and Yuan]{lin2023video}
Bin Lin, Yang Ye, Bin Zhu, Jiaxi Cui, Munan Ning, Peng Jin, and Li Yuan.
\newblock Video-llava: Learning united visual representation by alignment before projection.
\newblock \emph{arXiv preprint arXiv:2311.10122}, 2023{\natexlab{a}}.

\bibitem[Lin et~al.(2024{\natexlab{a}})Lin, Yin, Ping, Molchanov, Shoeybi, and Han]{lin2024vila}
Ji Lin, Hongxu Yin, Wei Ping, Pavlo Molchanov, Mohammad Shoeybi, and Song Han.
\newblock Vila: On pre-training for visual language models.
\newblock In \emph{Proceedings of the IEEE/CVF Conference on Computer Vision and Pattern Recognition}, pages 26689--26699, 2024{\natexlab{a}}.

\bibitem[Lin et~al.(2014)Lin, Maire, Belongie, Hays, Perona, Ramanan, Doll{\'{a}}r, and Zitnick]{Datasets:MSCOCO}
Tsung{-}Yi Lin, Michael Maire, Serge~J. Belongie, James Hays, Pietro Perona, Deva Ramanan, Piotr Doll{\'{a}}r, and C.~Lawrence Zitnick.
\newblock Microsoft {COCO:} common objects in context.
\newblock In \emph{ECCV}, pages 740--755, 2014.

\bibitem[Lin et~al.(2024{\natexlab{b}})Lin, Wei, An, Gao, Zou, Luo, Huang, Zhang, and Li]{lin2024draw}
Weifeng Lin, Xinyu Wei, Ruichuan An, Peng Gao, Bocheng Zou, Yulin Luo, Siyuan Huang, Shanghang Zhang, and Hongsheng Li.
\newblock Draw-and-understand: Leveraging visual prompts to enable mllms to comprehend what you want.
\newblock \emph{arXiv preprint arXiv:2403.20271}, 2024{\natexlab{b}}.

\bibitem[Lin et~al.(2023{\natexlab{b}})Lin, Liu, Zhang, Gao, Qiu, Xiao, Qiu, Lin, Shao, Chen, et~al.]{lin2023sphinx}
Ziyi Lin, Chris Liu, Renrui Zhang, Peng Gao, Longtian Qiu, Han Xiao, Han Qiu, Chen Lin, Wenqi Shao, Keqin Chen, et~al.
\newblock Sphinx: The joint mixing of weights, tasks, and visual embeddings for multi-modal large language models.
\newblock \emph{arXiv preprint arXiv:2311.07575}, 2023{\natexlab{b}}.

\bibitem[Liu et~al.(2023{\natexlab{a}})Liu, Li, Li, and Lee]{VLM:LLaVA-1.5}
Haotian Liu, Chunyuan Li, Yuheng Li, and Yong~Jae Lee.
\newblock Improved baselines with visual instruction tuning.
\newblock \emph{arXiv: 2310.03744}, 2023{\natexlab{a}}.

\bibitem[Liu et~al.(2023{\natexlab{b}})Liu, Li, Wu, and Lee]{VLM:LLaVA}
Haotian Liu, Chunyuan Li, Qingyang Wu, and Yong~Jae Lee.
\newblock Visual instruction tuning.
\newblock In \emph{NeurIPS}, 2023{\natexlab{b}}.

\bibitem[Liu et~al.(2024{\natexlab{a}})Liu, Li, Li, Li, Zhang, Shen, and Lee]{VLM:LLaVA-1.6}
Haotian Liu, Chunyuan Li, Yuheng Li, Bo Li, Yuanhan Zhang, Sheng Shen, and Yong~Jae Lee.
\newblock Llava-next: Improved reasoning, ocr, and world knowledge, 2024{\natexlab{a}}.

\bibitem[Liu et~al.(2023{\natexlab{c}})Liu, Zeng, Ren, Li, Zhang, Yang, Jiang, Li, Yang, Su, et~al.]{liu2023grounding}
Shilong Liu, Zhaoyang Zeng, Tianhe Ren, Feng Li, Hao Zhang, Jie Yang, Qing Jiang, Chunyuan Li, Jianwei Yang, Hang Su, et~al.
\newblock Grounding dino: Marrying dino with grounded pre-training for open-set object detection.
\newblock \emph{arXiv preprint arXiv:2303.05499}, 2023{\natexlab{c}}.

\bibitem[Liu et~al.(2023{\natexlab{d}})Liu, Duan, Zhang, Li, Zhang, Zhao, Yuan, Wang, He, Liu, Chen, and Lin]{Datasets:MMBench}
Yuan Liu, Haodong Duan, Yuanhan Zhang, Bo Li, Songyang Zhang, Wangbo Zhao, Yike Yuan, Jiaqi Wang, Conghui He, Ziwei Liu, Kai Chen, and Dahua Lin.
\newblock Mmbench: Is your multi-modal model an all-around player?
\newblock \emph{arXiv: 2307.06281}, 2023{\natexlab{d}}.

\bibitem[Liu et~al.(2024{\natexlab{b}})Liu, Li, Huang, Yang, Yu, Li, Yin, Liu, Jin, and Bai]{liu2024ocrbench}
Yuliang Liu, Zhang Li, Mingxin Huang, Biao Yang, Wenwen Yu, Chunyuan Li, Xu-Cheng Yin, Cheng-Lin Liu, Lianwen Jin, and Xiang Bai.
\newblock Ocrbench: on the hidden mystery of ocr in large multimodal models.
\newblock \emph{Science China Information Sciences}, 67\penalty0 (12):\penalty0 220102, 2024{\natexlab{b}}.

\bibitem[Liu et~al.(2021)Liu, Lin, Cao, Hu, Wei, Zhang, Lin, and Guo]{liu2021swin}
Ze Liu, Yutong Lin, Yue Cao, Han Hu, Yixuan Wei, Zheng Zhang, Stephen Lin, and Baining Guo.
\newblock Swin transformer: Hierarchical vision transformer using shifted windows.
\newblock In \emph{ICCV}, pages 10012--10022, 2021.

\bibitem[Liu et~al.(2022)Liu, Mao, Wu, Feichtenhofer, Darrell, and Xie]{liu2022convnet}
Zhuang Liu, Hanzi Mao, Chao-Yuan Wu, Christoph Feichtenhofer, Trevor Darrell, and Saining Xie.
\newblock A convnet for the 2020s.
\newblock In \emph{Proceedings of the IEEE/CVF conference on computer vision and pattern recognition}, pages 11976--11986, 2022.

\bibitem[Long et~al.(2022)Long, Qin, Panteleev, Bissacco, Fujii, and Raptis]{hiertext}
Shangbang Long, Siyang Qin, Dmitry Panteleev, Alessandro Bissacco, Yasuhisa Fujii, and Michalis Raptis.
\newblock Towards end-to-end unified scene text detection and layout analysis.
\newblock In \emph{Proceedings of the IEEE/CVF Conference on Computer Vision and Pattern Recognition}, pages 1049--1059, 2022.

\bibitem[Luo et~al.(2024{\natexlab{a}})Luo, Zhou, Zhang, Zheng, Sun, and Ji]{luo2024feast}
Gen Luo, Yiyi Zhou, Yuxin Zhang, Xiawu Zheng, Xiaoshuai Sun, and Rongrong Ji.
\newblock Feast your eyes: Mixture-of-resolution adaptation for multimodal large language models.
\newblock \emph{arXiv preprint arXiv:2403.03003}, 2024{\natexlab{a}}.

\bibitem[Luo et~al.(2024{\natexlab{b}})Luo, Zhou, Zhang, Zheng, Sun, and Ji]{luo2024llava_hr}
Gen Luo, Yiyi Zhou, Yuxin Zhang, Xiawu Zheng, Xiaoshuai Sun, and Rongrong Ji.
\newblock Feast your eyes: Mixture-of-resolution adaptation for multimodal large language models.
\newblock \emph{arXiv preprint arXiv:2403.03003}, 2024{\natexlab{b}}.

\bibitem[Lv et~al.(2023)Lv, Huang, Chen, Cui, Ma, Chang, Huang, Wang, Dong, Luo, et~al.]{lv2023kosmos2_5}
Tengchao Lv, Yupan Huang, Jingye Chen, Lei Cui, Shuming Ma, Yaoyao Chang, Shaohan Huang, Wenhui Wang, Li Dong, Weiyao Luo, et~al.
\newblock Kosmos-2.5: A multimodal literate model.
\newblock \emph{arXiv preprint arXiv:2309.11419}, 2023.

\bibitem[Ma et~al.(2024)Ma, Jiang, Wu, Yuan, and Qi]{ma2024groma}
Chuofan Ma, Yi Jiang, Jiannan Wu, Zehuan Yuan, and Xiaojuan Qi.
\newblock Groma: Localized visual tokenization for grounding multimodal large language models.
\newblock \emph{arXiv preprint arXiv:2404.13013}, 2024.

\bibitem[Mao et~al.(2016)Mao, Huang, Toshev, Camburu, Yuille, and Murphy]{mao2016refcocog}
Junhua Mao, Jonathan Huang, Alexander Toshev, Oana Camburu, Alan~L Yuille, and Kevin Murphy.
\newblock Generation and comprehension of unambiguous object descriptions.
\newblock In \emph{CVPR}, pages 11--20, 2016.

\bibitem[McKinzie et~al.(2024)McKinzie, Gan, Fauconnier, Dodge, Zhang, Dufter, Shah, Du, Peng, Weers, Belyi, Zhang, Singh, Kang, Jain, H{\`{e}}, Schwarzer, Gunter, Kong, Zhang, Wang, Wang, Du, Lei, Wiseman, Yin, Lee, Wang, Pang, Grasch, Toshev, and Yang]{VLM:MM1}
Brandon McKinzie, Zhe Gan, Jean{-}Philippe Fauconnier, Sam Dodge, Bowen Zhang, Philipp Dufter, Dhruti Shah, Xianzhi Du, Futang Peng, Floris Weers, Anton Belyi, Haotian Zhang, Karanjeet Singh, Doug Kang, Ankur Jain, Hongyu H{\`{e}}, Max Schwarzer, Tom Gunter, Xiang Kong, Aonan Zhang, Jianyu Wang, Chong Wang, Nan Du, Tao Lei, Sam Wiseman, Guoli Yin, Mark Lee, Zirui Wang, Ruoming Pang, Peter Grasch, Alexander Toshev, and Yinfei Yang.
\newblock {MM1:} methods, analysis {\&} insights from multimodal {LLM} pre-training.
\newblock \emph{arXiv: 2403.09611}, 2024.

\bibitem[OpenAI(2023)]{gpt4v}
OpenAI.
\newblock Gpt-4v(ision) system card.
\newblock \url{https://cdn.openai.com/papers/GPTV_System_Card.pdf}, 2023.

\bibitem[Peng et~al.(2023)Peng, Wang, Dong, Hao, Huang, Ma, and Wei]{peng2023kosmos2}
Zhiliang Peng, Wenhui Wang, Li Dong, Yaru Hao, Shaohan Huang, Shuming Ma, and Furu Wei.
\newblock Kosmos-2: Grounding multimodal large language models to the world.
\newblock \emph{arXiv preprint arXiv:2306.14824}, 2023.

\bibitem[Pi et~al.(2024)Pi, Yao, Gao, Zhang, and Zhang]{pi2024perceptiongpt}
Renjie Pi, Lewei Yao, Jiahui Gao, Jipeng Zhang, and Tong Zhang.
\newblock Perceptiongpt: Effectively fusing visual perception into llm.
\newblock In \emph{Proceedings of the IEEE/CVF Conference on Computer Vision and Pattern Recognition}, pages 27124--27133, 2024.

\bibitem[Radford et~al.(2021)Radford, Kim, Hallacy, Ramesh, Goh, Agarwal, Sastry, Askell, Mishkin, Clark, Krueger, and Sutskever]{VLP:CLIP}
Alec Radford, Jong~Wook Kim, Chris Hallacy, Aditya Ramesh, Gabriel Goh, Sandhini Agarwal, Girish Sastry, Amanda Askell, Pamela Mishkin, Jack Clark, Gretchen Krueger, and Ilya Sutskever.
\newblock Learning transferable visual models from natural language supervision.
\newblock In \emph{ICML}, pages 8748--8763, 2021.

\bibitem[Ramanathan et~al.(2023)Ramanathan, Kalia, Petrovic, Wen, Zheng, Guo, Wang, Marquez, Kovvuri, Kadian, et~al.]{ramanathan2023paco}
Vignesh Ramanathan, Anmol Kalia, Vladan Petrovic, Yi Wen, Baixue Zheng, Baishan Guo, Rui Wang, Aaron Marquez, Rama Kovvuri, Abhishek Kadian, et~al.
\newblock Paco: Parts and attributes of common objects.
\newblock In \emph{Proceedings of the IEEE/CVF Conference on Computer Vision and Pattern Recognition}, pages 7141--7151, 2023.

\bibitem[Rasheed et~al.(2024)Rasheed, Maaz, Shaji, Shaker, Khan, Cholakkal, Anwer, Xing, Yang, and Khan]{rasheed2024glamm}
Hanoona Rasheed, Muhammad Maaz, Sahal Shaji, Abdelrahman Shaker, Salman Khan, Hisham Cholakkal, Rao~M Anwer, Eric Xing, Ming-Hsuan Yang, and Fahad~S Khan.
\newblock Glamm: Pixel grounding large multimodal model.
\newblock In \emph{Proceedings of the IEEE/CVF Conference on Computer Vision and Pattern Recognition}, pages 13009--13018, 2024.

\bibitem[Ren et~al.(2015)Ren, He, Girshick, and Sun]{Detection:FasterR-CNN}
Shaoqing Ren, Kaiming He, Ross~B. Girshick, and Jian Sun.
\newblock Faster {R-CNN:} towards real-time object detection with region proposal networks.
\newblock In \emph{NIPS}, pages 91--99, 2015.

\bibitem[Ren et~al.(2024)Ren, Jiang, Liu, Zeng, Liu, Gao, Huang, Ma, Jiang, Chen, et~al.]{ren2024grounding}
Tianhe Ren, Qing Jiang, Shilong Liu, Zhaoyang Zeng, Wenlong Liu, Han Gao, Hongjie Huang, Zhengyu Ma, Xiaoke Jiang, Yihao Chen, et~al.
\newblock Grounding dino 1.5: Advance the" edge" of open-set object detection.
\newblock \emph{arXiv preprint arXiv:2405.10300}, 2024.

\bibitem[Rezatofighi et~al.(2019)Rezatofighi, Tsoi, Gwak, Sadeghian, Reid, and Savarese]{rezatofighi2019generalized}
Hamid Rezatofighi, Nathan Tsoi, JunYoung Gwak, Amir Sadeghian, Ian Reid, and Silvio Savarese.
\newblock Generalized intersection over union: A metric and a loss for bounding box regression.
\newblock In \emph{Proceedings of the IEEE/CVF conference on computer vision and pattern recognition}, pages 658--666, 2019.

\bibitem[Schuhmann et~al.(2022)Schuhmann, Beaumont, Vencu, Gordon, Wightman, Cherti, Coombes, Katta, Mullis, Wortsman, et~al.]{Datasets:Laion-5b}
Christoph Schuhmann, Romain Beaumont, Richard Vencu, Cade Gordon, Ross Wightman, Mehdi Cherti, Theo Coombes, Aarush Katta, Clayton Mullis, Mitchell Wortsman, et~al.
\newblock Laion-5b: An open large-scale dataset for training next generation image-text models.
\newblock \emph{NeurIPS}, 35:\penalty0 25278--25294, 2022.

\bibitem[Shao et~al.(2018)Shao, Zhao, Li, Xiao, Yu, Zhang, and Sun]{shao2018crowdhuman}
Shuai Shao, Zijian Zhao, Boxun Li, Tete Xiao, Gang Yu, Xiangyu Zhang, and Jian Sun.
\newblock Crowdhuman: A benchmark for detecting human in a crowd.
\newblock \emph{arXiv preprint arXiv:1805.00123}, 2018.

\bibitem[Shao et~al.(2019)Shao, Li, Zhang, Peng, Yu, Zhang, Li, and Sun]{shao2019objects365}
Shuai Shao, Zeming Li, Tianyuan Zhang, Chao Peng, Gang Yu, Xiangyu Zhang, Jing Li, and Jian Sun.
\newblock Objects365: A large-scale, high-quality dataset for object detection.
\newblock In \emph{ICCV}, pages 8430--8439, 2019.

\bibitem[Shi et~al.(2024)Shi, Liu, Wang, Liao, Radhakrishnan, Huang, Yin, Sapra, Yacoob, Shi, et~al.]{shi2024eagle}
Min Shi, Fuxiao Liu, Shihao Wang, Shijia Liao, Subhashree Radhakrishnan, De-An Huang, Hongxu Yin, Karan Sapra, Yaser Yacoob, Humphrey Shi, et~al.
\newblock Eagle: Exploring the design space for multimodal llms with mixture of encoders.
\newblock \emph{arXiv preprint arXiv:2408.15998}, 2024.

\bibitem[Singh et~al.(2019)Singh, Natarajan, Shah, Jiang, Chen, Batra, Parikh, and Rohrbach]{singh2019towards}
Amanpreet Singh, Vivek Natarajan, Meet Shah, Yu Jiang, Xinlei Chen, Dhruv Batra, Devi Parikh, and Marcus Rohrbach.
\newblock Towards vqa models that can read.
\newblock In \emph{Proceedings of the IEEE/CVF conference on computer vision and pattern recognition}, pages 8317--8326, 2019.

\bibitem[Taori et~al.(2023)Taori, Gulrajani, Zhang, Dubois, Li, Guestrin, Liang, and Hashimoto]{TransF:Alpaca}
Rohan Taori, Ishaan Gulrajani, Tianyi Zhang, Yann Dubois, Xuechen Li, Carlos Guestrin, Percy Liang, and Tatsunori~B. Hashimoto.
\newblock Stanford alpaca: An instruction-following llama model.
\newblock \url{https://github.com/tatsu-lab/stanford_alpaca}, 2023.

\bibitem[Team et~al.(2023)Team, Anil, Borgeaud, Wu, Alayrac, Yu, Soricut, Schalkwyk, Dai, Hauth, et~al.]{VLM:Gemini}
Gemini Team, Rohan Anil, Sebastian Borgeaud, Yonghui Wu, Jean-Baptiste Alayrac, Jiahui Yu, Radu Soricut, Johan Schalkwyk, Andrew~M Dai, Anja Hauth, et~al.
\newblock Gemini: a family of highly capable multimodal models.
\newblock \emph{arXiv: 2312.11805}, 2023.

\bibitem[Team(2023)]{TransF:InternLM}
InternLM Team.
\newblock Internlm: A multilingual language model with progressively enhanced capabilities.
\newblock \url{https://github.com/InternLM/InternLM}, 2023.

\bibitem[Tong et~al.(2025)Tong, Brown, Wu, Woo, IYER, Akula, Yang, Yang, Middepogu, Wang, et~al.]{tong2025cambrian}
Peter Tong, Ellis Brown, Penghao Wu, Sanghyun Woo, Adithya Jairam~Vedagiri IYER, Sai~Charitha Akula, Shusheng Yang, Jihan Yang, Manoj Middepogu, Ziteng Wang, et~al.
\newblock Cambrian-1: A fully open, vision-centric exploration of multimodal llms.
\newblock \emph{Advances in Neural Information Processing Systems}, 37:\penalty0 87310--87356, 2025.

\bibitem[Tong et~al.(2024)Tong, Brown, Wu, Woo, Middepogu, Akula, Yang, Yang, Iyer, Pan, et~al.]{tong2024cambrian}
Shengbang Tong, Ellis Brown, Penghao Wu, Sanghyun Woo, Manoj Middepogu, Sai~Charitha Akula, Jihan Yang, Shusheng Yang, Adithya Iyer, Xichen Pan, et~al.
\newblock Cambrian-1: A fully open, vision-centric exploration of multimodal llms.
\newblock \emph{arXiv preprint arXiv:2406.16860}, 2024.

\bibitem[Touvron et~al.(2023{\natexlab{a}})Touvron, Lavril, Izacard, Martinet, Lachaux, Lacroix, Rozi{\`{e}}re, Goyal, Hambro, Azhar, Rodriguez, Joulin, Grave, and Lample]{TransF:LLaMA}
Hugo Touvron, Thibaut Lavril, Gautier Izacard, Xavier Martinet, Marie{-}Anne Lachaux, Timoth{\'{e}}e Lacroix, Baptiste Rozi{\`{e}}re, Naman Goyal, Eric Hambro, Faisal Azhar, Aur{\'{e}}lien Rodriguez, Armand Joulin, Edouard Grave, and Guillaume Lample.
\newblock Llama: Open and efficient foundation language models.
\newblock \emph{arXiv: 2302.13971}, 2023{\natexlab{a}}.

\bibitem[Touvron et~al.(2023{\natexlab{b}})Touvron, Martin, Stone, Albert, Almahairi, Babaei, Bashlykov, Batra, Bhargava, Bhosale, et~al.]{TransF:LLaMA2}
Hugo Touvron, Louis Martin, Kevin Stone, Peter Albert, Amjad Almahairi, Yasmine Babaei, Nikolay Bashlykov, Soumya Batra, Prajjwal Bhargava, Shruti Bhosale, et~al.
\newblock Llama 2: Open foundation and fine-tuned chat models.
\newblock \emph{arXiv: 2307.09288}, 2023{\natexlab{b}}.

\bibitem[Wang et~al.(2024)Wang, Bai, Tan, Wang, Fan, Bai, Chen, Liu, Wang, Ge, et~al.]{wang2024qwen2}
Peng Wang, Shuai Bai, Sinan Tan, Shijie Wang, Zhihao Fan, Jinze Bai, Keqin Chen, Xuejing Liu, Jialin Wang, Wenbin Ge, et~al.
\newblock Qwen2-vl: Enhancing vision-language model's perception of the world at any resolution.
\newblock \emph{arXiv preprint arXiv:2409.12191}, 2024.

\bibitem[Wang et~al.(2023)Wang, Lv, Yu, Hong, Qi, Wang, Ji, Yang, Zhao, Song, et~al.]{wang2023cogvlm}
Weihan Wang, Qingsong Lv, Wenmeng Yu, Wenyi Hong, Ji Qi, Yan Wang, Junhui Ji, Zhuoyi Yang, Lei Zhao, Xixuan Song, et~al.
\newblock Cogvlm: Visual expert for pretrained language models.
\newblock \emph{arXiv preprint arXiv:2311.03079}, 2023.

\bibitem[Wu et~al.(2024)Wu, Zhong, Xing, Lai, Liu, Wang, Chen, Zhu, Lu, Lu, et~al.]{wu2024visionllm}
Jiannan Wu, Muyan Zhong, Sen Xing, Zeqiang Lai, Zhaoyang Liu, Wenhai Wang, Zhe Chen, Xizhou Zhu, Lewei Lu, Tong Lu, et~al.
\newblock Visionllm v2: An end-to-end generalist multimodal large language model for hundreds of vision-language tasks.
\newblock \emph{arXiv preprint arXiv:2406.08394}, 2024.

\bibitem[Xiao et~al.(2024)Xiao, Wu, Xu, Dai, Hu, Lu, Zeng, Liu, and Yuan]{xiao2024florence}
Bin Xiao, Haiping Wu, Weijian Xu, Xiyang Dai, Houdong Hu, Yumao Lu, Michael Zeng, Ce Liu, and Lu Yuan.
\newblock Florence-2: Advancing a unified representation for a variety of vision tasks.
\newblock In \emph{Proceedings of the IEEE/CVF Conference on Computer Vision and Pattern Recognition}, pages 4818--4829, 2024.

\bibitem[Xu et~al.(2024)Xu, Yao, Guo, Cui, Ni, Ge, Chua, Liu, Sun, and Huang]{VLM:LLaVA-UHD}
Ruyi Xu, Yuan Yao, Zonghao Guo, Junbo Cui, Zanlin Ni, Chunjiang Ge, Tat{-}Seng Chua, Zhiyuan Liu, Maosong Sun, and Gao Huang.
\newblock Llava-uhd: an {LMM} perceiving any aspect ratio and high-resolution images.
\newblock \emph{arXiv: 2403.11703}, 2024.

\bibitem[Xue et~al.(2024{\natexlab{a}})Xue, Chen, Li, Hu, Zhu, Li, Fang, Tang, Yang, Liu, et~al.]{xue2024longvila}
Fuzhao Xue, Yukang Chen, Dacheng Li, Qinghao Hu, Ligeng Zhu, Xiuyu Li, Yunhao Fang, Haotian Tang, Shang Yang, Zhijian Liu, et~al.
\newblock Longvila: Scaling long-context visual language models for long videos.
\newblock \emph{arXiv preprint arXiv:2408.10188}, 2024{\natexlab{a}}.

\bibitem[Xue et~al.(2024{\natexlab{b}})Xue, Shu, Awadalla, Wang, Yan, Purushwalkam, Zhou, Prabhu, Dai, Ryoo, et~al.]{xue2024xgen}
Le Xue, Manli Shu, Anas Awadalla, Jun Wang, An Yan, Senthil Purushwalkam, Honglu Zhou, Viraj Prabhu, Yutong Dai, Michael~S Ryoo, et~al.
\newblock xgen-mm (blip-3): A family of open large multimodal models.
\newblock \emph{arXiv preprint arXiv:2408.08872}, 2024{\natexlab{b}}.

\bibitem[Yang et~al.(2024)Yang, Yang, Zhang, Hui, Zheng, Yu, Li, Liu, Huang, Wei, et~al.]{yang2024qwen2}
An Yang, Baosong Yang, Beichen Zhang, Binyuan Hui, Bo Zheng, Bowen Yu, Chengyuan Li, Dayiheng Liu, Fei Huang, Haoran Wei, et~al.
\newblock Qwen2. 5 technical report.
\newblock \emph{arXiv preprint arXiv:2412.15115}, 2024.

\bibitem[Yang et~al.(2023)Yang, Zhang, Li, Zou, Li, and Gao]{yang2023set}
Jianwei Yang, Hao Zhang, Feng Li, Xueyan Zou, Chunyuan Li, and Jianfeng Gao.
\newblock Set-of-mark prompting unleashes extraordinary visual grounding in gpt-4v.
\newblock \emph{arXiv preprint arXiv:2310.11441}, 2023.

\bibitem[You et~al.(2023)You, Zhang, Gan, Du, Zhang, Wang, Cao, Chang, and Yang]{you2023ferret}
Haoxuan You, Haotian Zhang, Zhe Gan, Xianzhi Du, Bowen Zhang, Zirui Wang, Liangliang Cao, Shih-Fu Chang, and Yinfei Yang.
\newblock Ferret: Refer and ground anything anywhere at any granularity.
\newblock \emph{arXiv preprint arXiv:2310.07704}, 2023.

\bibitem[Yu et~al.(2016)Yu, Poirson, Yang, Berg, and Berg]{yu2016refcoco}
Licheng Yu, Patrick Poirson, Shan Yang, Alexander~C Berg, and Tamara~L Berg.
\newblock Modeling context in referring expressions.
\newblock In \emph{ECCV}, pages 69--85, 2016.

\bibitem[Yu et~al.(2023)Yu, Yang, Li, Wang, Lin, Liu, Wang, and Wang]{yu2023mmvet}
Weihao Yu, Zhengyuan Yang, Linjie Li, Jianfeng Wang, Kevin Lin, Zicheng Liu, Xinchao Wang, and Lijuan Wang.
\newblock Mm-vet: Evaluating large multimodal models for integrated capabilities.
\newblock \emph{arXiv preprint arXiv:2308.02490}, 2023.

\bibitem[Yuan et~al.(2024)Yuan, Li, Liu, Tang, Luo, Qin, Zhang, and Zhu]{yuan2024osprey}
Yuqian Yuan, Wentong Li, Jian Liu, Dongqi Tang, Xinjie Luo, Chi Qin, Lei Zhang, and Jianke Zhu.
\newblock Osprey: Pixel understanding with visual instruction tuning.
\newblock In \emph{Proceedings of the IEEE/CVF Conference on Computer Vision and Pattern Recognition}, pages 28202--28211, 2024.

\bibitem[Yue et~al.(2023)Yue, Ni, Zhang, Zheng, Liu, Zhang, Stevens, Jiang, Ren, Sun, et~al.]{yue2023mmmu}
Xiang Yue, Yuansheng Ni, Kai Zhang, Tianyu Zheng, Ruoqi Liu, Ge Zhang, Samuel Stevens, Dongfu Jiang, Weiming Ren, Yuxuan Sun, et~al.
\newblock Mmmu: A massive multi-discipline multimodal understanding and reasoning benchmark for expert agi.
\newblock \emph{arXiv preprint arXiv:2311.16502}, 2023.

\bibitem[Zellers et~al.(2019)Zellers, Bisk, Farhadi, and Choi]{zellers2019vcr}
Rowan Zellers, Yonatan Bisk, Ali Farhadi, and Yejin Choi.
\newblock From recognition to cognition: Visual commonsense reasoning.
\newblock In \emph{The IEEE Conference on Computer Vision and Pattern Recognition (CVPR)}, 2019.

\bibitem[Zhan et~al.(2024)Zhan, Zhu, Zhao, Yang, Tang, and Wang]{zhan2024griffon}
Yufei Zhan, Yousong Zhu, Hongyin Zhao, Fan Yang, Ming Tang, and Jinqiao Wang.
\newblock Griffon v2: Advancing multimodal perception with high-resolution scaling and visual-language co-referring.
\newblock \emph{arXiv preprint arXiv:2403.09333}, 2024.

\bibitem[Zhan et~al.(2025)Zhan, Zhu, Chen, Yang, Tang, and Wang]{zhan2025griffon}
Yufei Zhan, Yousong Zhu, Zhiyang Chen, Fan Yang, Ming Tang, and Jinqiao Wang.
\newblock Griffon: Spelling out all object locations at any granularity with large language models.
\newblock In \emph{European Conference on Computer Vision}, pages 405--422. Springer, 2025.

\bibitem[Zhang et~al.(2022{\natexlab{a}})Zhang, Li, Liu, Zhang, Su, Zhu, Ni, and Shum]{zhang2022dino}
Hao Zhang, Feng Li, Shilong Liu, Lei Zhang, Hang Su, Jun Zhu, Lionel~M Ni, and Heung-Yeung Shum.
\newblock Dino: Detr with improved denoising anchor boxes for end-to-end object detection.
\newblock \emph{arXiv preprint arXiv:2203.03605}, 2022{\natexlab{a}}.

\bibitem[Zhang et~al.(2024{\natexlab{a}})Zhang, Gao, Gan, Dufter, Wenzel, Huang, Shah, Du, Zhang, Li, et~al.]{zhang2024mm1}
Haotian Zhang, Mingfei Gao, Zhe Gan, Philipp Dufter, Nina Wenzel, Forrest Huang, Dhruti Shah, Xianzhi Du, Bowen Zhang, Yanghao Li, et~al.
\newblock Mm1. 5: Methods, analysis \& insights from multimodal llm fine-tuning.
\newblock \emph{arXiv preprint arXiv:2409.20566}, 2024{\natexlab{a}}.

\bibitem[Zhang et~al.(2024{\natexlab{b}})Zhang, You, Dufter, Zhang, Chen, Chen, Fu, Wang, Chang, Gan, et~al.]{zhang2024ferret}
Haotian Zhang, Haoxuan You, Philipp Dufter, Bowen Zhang, Chen Chen, Hong-You Chen, Tsu-Jui Fu, William~Yang Wang, Shih-Fu Chang, Zhe Gan, et~al.
\newblock Ferret-v2: An improved baseline for referring and grounding with large language models.
\newblock \emph{arXiv preprint arXiv:2404.07973}, 2024{\natexlab{b}}.

\bibitem[Zhang et~al.(2025)Zhang, Li, Li, Ren, Zou, Liu, Huang, Gao, Li, Yang, et~al.]{zhang2025llava}
Hao Zhang, Hongyang Li, Feng Li, Tianhe Ren, Xueyan Zou, Shilong Liu, Shijia Huang, Jianfeng Gao, Chunyuan Li, Jainwei Yang, et~al.
\newblock Llava-grounding: Grounded visual chat with large multimodal models.
\newblock In \emph{European Conference on Computer Vision}, pages 19--35. Springer, 2025.

\bibitem[Zhang et~al.(2023)Zhang, Sun, Chen, Xiao, Shao, Zhang, Chen, and Luo]{VLM:GPT4ROI}
Shilong Zhang, Peize Sun, Shoufa Chen, Min Xiao, Wenqi Shao, Wenwei Zhang, Kai Chen, and Ping Luo.
\newblock Gpt4roi: Instruction tuning large language model on region-of-interest.
\newblock \emph{arXiv: 2307.03601}, 2023.

\bibitem[Zhang et~al.(2022{\natexlab{b}})Zhang, Sun, Zhou, He, Yin, Wang, Sheng, Qiao, Shao, and Liu]{zhang2022bamboo}
Yuanhan Zhang, Qinghong Sun, Yichun Zhou, Zexin He, Zhenfei Yin, Kun Wang, Lu Sheng, Yu Qiao, Jing Shao, and Ziwei Liu.
\newblock Bamboo: Building mega-scale vision dataset continually with human-machine synergy.
\newblock \emph{arXiv preprint arXiv:2203.07845}, 2022{\natexlab{b}}.

\bibitem[Zhao et~al.(2024)Zhao, Li, Duan, Huang, Li, Chen, and Yang]{zhao2024mg}
Xiangyu Zhao, Xiangtai Li, Haodong Duan, Haian Huang, Yining Li, Kai Chen, and Hua Yang.
\newblock Mg-llava: Towards multi-granularity visual instruction tuning.
\newblock \emph{arXiv preprint arXiv:2406.17770}, 2024.

\bibitem[Zhu et~al.(2023)Zhu, Xiao, Alvarado, Babaei, Hu, El-Mohri, Culatana, Sumbaly, and Yan]{zhu2023egoobjects}
Chenchen Zhu, Fanyi Xiao, Andr{\'e}s Alvarado, Yasmine Babaei, Jiabo Hu, Hichem El-Mohri, Sean Culatana, Roshan Sumbaly, and Zhicheng Yan.
\newblock Egoobjects: A large-scale egocentric dataset for fine-grained object understanding.
\newblock In \emph{Proceedings of the IEEE/CVF International Conference on Computer Vision}, pages 20110--20120, 2023.

\end{thebibliography}
